\documentclass[smallextended]{svjour3}       
\smartqed  
\usepackage{graphicx}
\usepackage{subcaption}
\usepackage{color}
\usepackage{hyperref}
\usepackage{amsmath}
\usepackage{amsfonts}
\usepackage{placeins}
\usepackage{algorithmicx}
\usepackage{algorithm}
\usepackage{algpseudocode}
\usepackage{booktabs}
\usepackage{tabularx}
\usepackage{multirow,makecell}
\usepackage{todonotes}

\newcolumntype{L}[1]{>{\raggedright\let\newline\\\arraybackslash\hspace{0pt}}m{#1}}

\newcommand\blue[1]{{\color{black}#1}} 
\algdef{SE}[DOWHILE]{Do}{doWhile}{\algorithmicdo}[1]{\algorithmicwhile\ #1}%

\journalname{Annals of Mathematics and Artificial Intelligence}
\begin{document}

\title{Soft computing methods for multiobjective location of garbage accumulation points in smart cities}

\titlerunning{Soft computing for multiobjective location of garbage accumulation points}

\author{Jamal Toutouh \and Diego Rossit \and Sergio Nesmachnow}

\authorrunning{J. Toutouh, D. Rossit, S. Nesmachnow}

\institute{J. Toutouh \at Massachusetts Institute of Technology, United States\\
\email{toutouh@mit.edu}
\and 
D. Rossit \at DI, Universidad Nacional del Sur and CONICET, Argentina \\
\email{diego.rossit@uns.edu.ar}
\and 
S. Nesmachnow \at Universidad de la Rep\'ublica, Uruguay\\
\email{sergion@fing.edu.uy}
}

\date{Received: date / Accepted: date}

\maketitle

\begin{abstract}

This article describes the application of soft computing methods for solving the problem of locating garbage accumulation points in urban scenarios. This is a relevant problem in modern smart cities, in order to reduce negative environmental and social impacts in the waste management process, and also to optimize the available budget from the city administration to install waste bins. A specific problem model is presented, which accounts for reducing the investment costs, enhance the number of citizens served by the installed bins, and the accessibility to the system. A family of \blue{single- and multi-objective} heuristics based on the PageRank method and \blue{two} mutiobjective evolutionary algorithms are proposed. Experimental evaluation performed on real scenarios on the cities of Montevideo (Uruguay) and Bah\'ia Blanca (Argentina) demonstrates the effectiveness of the proposed approaches. The methods allow computing plannings with different trade-off between the problem objectives. The computed results improve over the current planning in Montevideo and provide a reasonable budget cost and quality of service for Bah\'ia Blanca.
\end{abstract}

\section{Introduction}
\label{Sec:intro}

The paradigm of smart cities proposes taking advantage of Information and Communication Technologies (ICTs) to manage assets and resources efficiently, in order to improve the quality and efficiency of urban services~\cite{Deakin2012}. 
Usually information processing and intelligent computational methods are applied to address important problems such as urban mobility, energy and sustainability, water supply and sewerage, waste management and recycling, pollution and environment, surveillance and security, healthcare and medicine, and many others. This way, a smart city can be more prepared to respond to specific challenges that affect the quality of life of its citizens~\cite{Khatoun2016}.

Municipal Solid Waste (MSW) management is one of the capital problems in modern smart cities. MSW management refers to the process of collecting, treating, and disposing of solid material that is discarded by citizens. The current increase of garbage generation rate in urban areas~\cite{hoornweg2015} has led to initiatives for encouraging efficient and sustainable practices. In fact, MSW management is one of the main challenges for local governments in order to mitigate environmental and social impacts, especially in highly populated cities, and reduce budgetary expenses. A specific problem related to the MSW system in urban scenarios refers to find a proper location for {collection sites} or \textit{Garbage Accumulation Points} (GAPs), i.e., the specific places where community waste bins are to be installed in order citizens can deposit their garbage. This is a relevant problem to guarantee a good service to citizens, because a paltry spatial distribution of waste bins in the city may lead to not fulfilling the needs of residents and/or affecting the quality of the service (e.g., people must walk long distances for garbage disposal, or certain waste bins fill quickly while others remain empty). 

Finding appropriate locations for waste bins is a variation of the Facility Location Problem, which is proved to be NP-hard~\cite{Megiddo1982}. In this context, where exact optimization methods require long execution times for solving realistic instances, soft computing methods such as heuristics and metaheuristics~\cite{Nesmachnow2014} are viable options to find good-quality approximate solutions, allowing to analyze different configurations for waste bins and also different scenarios. Moreover, the aforementioned capabilities of soft computing methods are important when planning MSW management in nowadays smart cities, especially considering that the final location plan for GAPs usually considers several criteria (thus, the underlying optimization problem is multiobjective). Furthermore, location plans must fulfill a set of hard constraints, and it is also desirable that waste bin locations change periodically, in order to not disturb the same citizens for a long time. This last feature is closely related with the `not in my backyard' phenomenon of semi-obnoxious facilities, such as waste bins, which is widley known in the literature \cite{lindell1983close}. Despite knowing that GAPs must be located somewhere, citizens are reluctant to have a waste bin very close to their homes, since it is generally linked to undesirable aspects, e.g., bad smell, visual pollution, disturbing noises, and heavy traffic associated with collecting vehicles.

The problem model proposed in this article considers performing the GAP distribution along the city while taking into account different criteria: reduce the expenses of installing bins, facilitate the accessibility to the system by reducing the walking distance of the citizens to the bins and serve as many people as possible, maximizing the total amount of waste collected. Two soft computing approaches are applied to solve the GAP location problem: \blue{four single- and multi-objective} PageRank heuristics and \blue{two state-of-the-art} Multiobjective Evolutionary Algorithms (MOEA). PageRank provides a simple approach for planning \blue{by taking local optima decisions}, while the MOEAs allow exploring solutions that account garbage accumulation points configurations with different trade-off between the problem objectives. The experimental evaluation was performed on real scenarios on Montevideo, Uruguay, and Bah\'ia Blanca, Argentina. The results in Montevideo demonstrated that significant improvements are obtained over the current solution implemented in the city (up to 90\% in distance and 31\% in cost). In the case of Bah\'ia Blanca, a city that still uses a door-to-door system in which the collection vehicle visits every dwelling, the solutions can be used as a starting point to migrate to a community bins based system that has certain advantages over the current system~\cite{rossitX2018municipal} and it is under consideration of the local authorities.

This article extends our previous conference article `Computational intelligence for locating garbage accumulation points in urban scenarios'~\cite{toutouhX2018intelligence} presented at the \textit{Learning and Intelligent Optimization Conference LION 12}. The main contributions of this article include improvements in the proposed methods, \blue{a multiobjective version of the PageRank algorithm, a new MOEA}, an extended experimental evaluation that incorporates real scenarios from Bah\'ia Blanca
Argentina,
\blue{the study of multiobjective optimization metrics},
and a specific analysis of quality of service for the computed solutions.

The article is structured as follows. Section \ref{sec:problem} introduces the problem. The proposed methods for solving the problem are described in Section \ref{sec:methods}. The experimental evaluation is reported in Section \ref{sec:exp}. Finally, Section \ref{sec:conclusion} presents the main conclusions and formulates the main lines for future work.

\section{The GAPs location problem}
\label{sec:problem}

This section describes the GAPs location problem and a review of related work applying 
soft computing methods to solve different variants of the problem.

\subsection{Problem definition and model}
The problem addressed in this article aims at selecting the best locations of GAPs from a predefined set of potential places and determine the number and type of waste bins (indicated, hereafter, as ``bins'') that are to be installed in each chosen GAP, according to three different criteria: i) maximize the total amount of waste collected; ii) minimize the installation cost of bins; and iii) minimize the average distance between the garbage generators and the assigned bins as a metric of the Quality of Service (QoS) offered to the users. 

The proposed problem model incorporates several features from situations arising in real cities, including:
\begin{itemize}
\item i) the assignment between the set of generators and set of GAPs is not a one-to-one correspondence, i.e., a generator can be assigned to more than one GAP. This is a realistic characteristic since when a generator finds a GAP that is full, it is likely that it will deposit the rest of their garbage in a nearby GAP. Moreover, if it is considered that, as it is common in the related literature (and will be explained in more detail in Section~\ref{subsec:formulation}), a generator represents actually a ``group'' of generators the behaviour of them can not be expected to be uniform in the sense that all deposit their waste in the same GAP.
\item ii) a generator cannot be assigned to a GAP that is located beyond a maximum threshold distance. Although the total distance between generators and assigned GAPs is minimized, there will be generators that will have larger transport distances than others. It is reasonable to limit the maximum transport distance that any generator has to carry their waste in order to keep a standard of fairness among users.
\item iii) the objective of maximization of total amount of waste it is also a realistic feature if the community bins system ``is in competition'' with other collection systems and it is not expected to collect all the waste. In cities of developing countries, such as Bah\'ia Blanca, it is known that a certain amount of garbage is collected by scavengers~\cite{wilsonX2006role} and, thus, remains outside of the formal collection system. Another example are the specific campaigns that some cities carry out to collect a certain fraction of waste. The City Hall of Bah\'ia Blanca has a few special places along the city which are called “clean points” in which different bins are located to receive different fractions of waste. Although the use of clean points, which are usually located in busy places, is mainly for educational and promotional purposes to encourage recycling rather than part of a real large-scale collection system, they can reduce the total amount of waste collected by the traditional system. Therefore, the use of this objective can help to build different solutions corresponding to different estimations of the percentage of waste that is expected to be collected by alternative systems. 
\end{itemize}

The problem goals account for different objectives that takes into account the point of view of both citizens and city administrators. On the one hand, the QoS of the waste management and collection system is considered, by accounting for the total waste collected in the installed bins. The point of view of citizens is directly addressed, as the average distance between the garbage generators and the assigned bins is proposed to be minimized.
On the other hand, economic considerations are taken into account, as the minimization of installation costs is also proposed, which in turn is of importance for city administrators.
This objectives are formally defined in the mathematical formulation of the problem, which is presented in the next subsection.

\subsection{Mathematical formulation}
\label{subsec:formulation}

The mathematical formulation of the GAPs location problem applies a Mixed Integer Programming (MIP) model.

Lets consider the following elements:

\begin{itemize}
\addtolength{\itemsep}{2pt}
\item A set $I = \{i_1, \ldots, i_M\}$ of potential GAPs for bins. Each GAP $i$ has an available space $S_{i}$ for installing bins.
\item A set $P = \{p_1, \ldots, p_N\}$ of generators. Following a usual approach in the related literature, nearby generators are grouped in clusters, assuming a similar behavior between elements in each cluster. The amount of waste produced by generator $p$ (in volumetric units) is $b_{p}$. The distance from generator $p$ to GAP $i \in I$ is $d_{pi}$, and the maximum distance between any generator in $P$ and its assigned GAP (in meters) is $D$. 
\item A set $J = \{j_1, \ldots, j_H\}$ of bin types. Each type has a given purchase price $c_j$, capacity $C_j$, and required space for its installation $e_j$. The maximum number of bins of type $j$ available is $MB_{j}$.
\end{itemize}

The model is described in Equations~\ref{eq:1}-\ref{eq:12}.
The following variables are used: $t_{ji}$ is the number of bins of type $j$ installed in GAP $i$, $x_{pi}$ is 1 if generator $p$ is assigned to GAP $i$ and 0 otherwise, and $f_{pi}$ is the fraction of the waste produced by generator $p$ that is deposited in GAP $i$. 
%
\begin{align}
\small
\label{eq:1}
\max & \sum_
{p\in P, \text{ } i\in I}
f_{pi} \times b_{p} \\[-1pt]
\label{eq:2}
\min & \sum_
{p\in P, \text{ } i\in I}
d_{pi} \times f_{pi} \\[-1pt]
\label{eq:3}
\min & \sum_
{j \in J, \text{ } h \in H, \text{ } i \in I}
t_{jhi} \times c_{j} \\[-1pt]
\nonumber
\text{subject to} \\[-1pt]
\label{eq:4}
\sum_{i\in{}I}\left(f_{pi}\right) \leq{} & 1 & \forall{}\ p\in{}P & \\[-1pt]
\label{eq:5}
\sum_
{j\in{}J, \text{ } h\in{}H}
\left(t_{ji} e_{j}\right) \leq{} & {S}_i & \forall{}\ i\in{}I & \\[-1pt]
\label{eq:6}
\sum_{\substack{p\in{}P}}\left(b_{p} f_{pi}\right) \leq{} & \sum_{j\in{}J}\left({C}_j t_{ji}\right) & \forall{}\ i\in{}I & \\[-1pt]
\label{eq:8}
d_{pi} x_{pi}\leq{}& D & \forall{}\ p\in{}P,\ i\in{}I & \\[-1pt]
\label{eq:9}
f_{pi} \leq{} &  x_{pi} & \forall{}\ p\in{}P,\ i\in{}I \\[-1pt]
\label{eq:10}
0 \leq{} f_{pi} \leq{} & 1 & \forall{}\ p\in{}P,\ i\in{}I \\[-1pt]
\label{eq:11}
x_{pi}&\in{}\left[0,1\right] & \forall{}\ p\in{}P,\ i\in{}I &\\[-1pt]
\label{eq:12}
t_{ji}&\in{{\mathbb{Z}}_{0}^{+}} & \forall{}\ j\in{}J,\ i\in{}I
\end{align}

\newpage
Three objective functions are proposed: the total waste that generators are able to dispose (Eq.~\ref{eq:1}); the total distance between generators and the assigned GAPs, weighted according to the waste fraction that is deposited in each GAP (Eq.~\ref{eq:2}); and the total investment cost (Eq.~\ref{eq:3}). Regarding the problem constraints, the sum of relative waste fractions of a generator must be less than one (Eq.~\ref{eq:4}); the space occupied with bins in a GAP must not be larger than the available space of the GAP (Eq.~\ref{eq:5}); the waste volume assigned to one GAP must not be larger than the storage capacity installed in that GAP, i.e., $S_{i}$ (Eq.~\ref{eq:6}); 
the maximum distance between a generator and any assigned GAP must be smaller than the threshold $D$ (Eq.~\ref{eq:8}); variable $x_{pi}$ is one if and only if some of the waste produced by generator $p$ is deposited in GAP $i$ (Eq.~\ref{eq:9}); the continuous variable $f_{pi}$ is defined between zero and one (Eq.~\ref{eq:10}), variable $x_{pi}$ is binary (Eq.~\ref{eq:11}), and variable $t_{ji}$ is a non-negative integer (Eq.~\ref{eq:12}).

The model considers that each block in the city contains a number of potential GAPs for bins. Each GAP has a limited available space ($S_{i}$) that restricts the number of bins that can be installed. Conversely to other approaches in the literature, the model considers the possibility that a generator can deposit its waste in several different GAPs. This is a rather realistic feature that is highly probable to occur in everyday life when a generator finds a GAP that has all its bins full. Moreover, taking into account that generators are grouped in clusters, instead of considering standalone individuals, this feature allows expressing that not all generators of a given cluster shall deposit their garbage in the same GAP.

\subsection{Related work}
\label{subec:related_work}

Regarding previous bibliographic reviews about the studied topic, Purkayastha et al.~\cite{purkayastha2015collection} made an effort to review the different applications related to bins location problems, concluding that the bibliography is rather scarce compare to the potential benefits that smart solutions in this problem can produce. Goulart et al.~\cite{goulartX2017multi} reviewed multicriteria approaches in municipal solid waste (MSW), emphasizing that there are again few applications that take advantage of the potential of multiobjective optimization in this field.

\blue{Due to aforementioned NP-hard nature of the problem, it is not surprising that the majority of the works that have addressed the GAP location problem have used heuristic or metaheuristics approaches. Moreover, despite the existence of some works that used exact methods, these approaches in general fail to properly handle large scale real-world instances. For example, Ghiani et al.~\cite{ghianiX2012capacitated} had to perform a partition of the original instance, which as a whole was solved heuristically, in order to apply an exact approach and even optimal solutions were not found after 3600 seconds of execution. In another example, the exact algorithm proposed by Rossit et al.~\cite{rossitX2018municipal} was highly time consuming, even when applied to a single objective version of the GAP location problem, being unable to find optimal solutions after 4200 seconds of execution.}

Several articles have presented heuristics and soft computing methods for solving problems that are similar to the GAPs location problem. Bautista and Pereira~\cite{bautista2006modeling} modeled the GAPs location problem as a minimal set covering/maximum satisfiability problem, and proposed a genetic algorithm and a GRASP metaheuristic for solving real instances in Barcelona, Spain. Other authors have applied integral approaches to solve the bins location problem and the routing/collection problem simultaneously. For example, Chang and Wei \cite{chang2000siting} used a fuzzy evolutionary search to solve the problem for a scenario in Kaohsiung, Taiwan. The model considered the percentage of population served, the average walking distance between users and their assigned GAP, and the approximate length of the routes of collecting vehicles as objectives. Explicit costs were not taken into account. Hemmelmay et al.~\cite{hemmelmayrX2013models} introduced the Waste Bin Allocation and Routing Problem, which was solved applying different methodologies that combine sequential and simultaneous strategies; the allocation was solved either with an exact or an heuristic method, while the routing was solved using Variable Neighborhood Search. 

A similar problem was addressed by Ghiani et al.~\cite{ghianiX2012capacitated}, 
using
a constructive heuristic for solving large 
instances in Nard\`{o}, Italy, that cannot be properly handled by exact methods implemented in CPLEX. Later \cite{ghianiX2014impact}, the heuristic was modified to bound posterior routing costs, e.g., not allowing the installation in a same GAP of bins that require different type of collecting vehicles. Di Felice \cite{di2014integrationpilotcase} 
proposed a two-phase heuristic for the problem, for a real case in L\textsc{\char13}Aquila, Italy. The first phase solved the location of the GAPs thorough a constructive heuristic, while the second determined the quantity and size of bins needed for each GA,P according to the number of waste generators served by that GAP. A similar heuristic was applied by Boskovic and Jovici~\cite{boskovic2015fast} for Kragujevac, Serbia, using the ArcGIS Network Analyst. Since bins location is a problem that uses spatial information, other authors relied on Geographic Information Systems to gather and analyze data. For example, Valeo et al.~\cite{valeo1998location} used a constructive heuristic to establish the GAPs sequentially according to some priorities in order to cover an studied area in Dundas, Canada.

Our research contributes with a novel model for solving the GAPs location problem, and determining which bins are to be installed in each GAP, \blue{providing a valuable decision support tool for decision makers in this field. The new model allows that a generator can visit several bins, according to his/her convenience}. This is a realistic aspect that has not been previously explored in the related literature. \blue{Moreover, as aforementioned, the inclusion of the objective of maximizing the collected waste is a valuable feature in developing countries}. A mathematical formulation of the problem and two soft computing approaches, i.e., PageRank and MOEAs, are proposed. Experiments performed on real cases in Montevideo and Bah\'ia Blanca show the competitiveness of the model and the proposed soft computing algorithms. Furthermore, since Montevideo has already an community bins system, a comparison with the current distribution of bins in the scenarios of this city is performed. This comparison shows that the proposed methods, in many cases, have improved the present distribution in terms of both investment cost and QoS provided to the citizens.

\section{Soft computing methods for the GAP location problem}
\label{sec:methods}

This section describes the soft computing approaches developed for solving the GAP location problem: \blue{four} PageRank heuristics and \blue{two} MOEAs. 

\subsection{PageRank algorithms for GAPs location}
\label{Sec:pagerank}

PageRank is a well-known voting algorithm to compute the relevance of web pages in Internet taking into account inbound and outbound links~\cite{Langville2011}. It has been successfully used to solve location problems that can be mathematically defined over graphs in the field of smart cities. 

The usual approach when applying PageRank for smart city problems consists in modeling the road networks as weighted graphs considering road-traffic information, and applying the voting algorithm to rank the potential locations to install infrastructure
\cite{Brahim2014,massobrioX2017infrastructure}. 

Weighted PageRank is applied to a given directed graph $G=(V,E)$ defined by a set of vertices $V$ and a set of edges $E$. The algorithm starts by initializing the PageRank value of each vertex $v_i$ to a fixed value $d$, i.e., $PR^W(v_i)=d,\ \forall\ v_i \in V$. 
\blue{The value} $d$ is known as the \textit{dumping parameter} and its default value is 0.85. Then, an iterative process is performed until a stop condition is reached. The stop condition usually involves a maximum number of iterations performed or the algorithm stops when a convergence value is below a given threshold. $PR^W(v_i)$ is computed according to the expression in Equation~\ref{eq:pagerank}, where $In(v_i)$ is the set of vertices that point to it (\textit{predecessors}), and $Out(v_i)$ is the set of vertices that $v_i$ points to (\textit{successors}), and $w_{ij}$ is the weight that for the edge that connects $v_i$ and $v_j$. 
\begin{equation}
\label{eq:pagerank}
PR^W(v_i) = (1-d) + d \times \left( \sum_{v_j \in In(v_i)} w_{ij} \times \frac{PR^W(v_j)}{ \sum\limits_{v_k \in Out(v_j)} w_{jk}}\right)
\end{equation}

In the GAPs location problem model, information about waste generators and collection points is modeled as a fully connected weighted graph $G=(V,E)$. $G$ is defined by the set of waste generators $P$ and the set of edges $E$. The weight of each edge $w_{jk} = \frac{b_j + b_k}{d_{jk}}$ relates the impact of the waste generated in both generators and their distance.
Thus, the tentative locations of GAPs are ranked in a sorted vector $I^{PR}$ in which $i^{PR}_j, i^{PR}_k \in I$, $j<k \Leftrightarrow PR^W(i^{PR}_j) > PR^W(s^{PR}_k$). 

Once the GAPs are sorted in $I^{PR}$, a constructive heuristic is applied to select a collection point configuration and locate it. 
The heuristic iterates over the sorted vector $I^{PR}$ starting by the best ranked element ($i^{PR}_1$). For each collection point $i^{PR}_j \in I^{PR}$, each of the three metrics evaluated 
(volume of the waste collected, distance from the generators to the assigned collection points, and installation costs) are evaluated for each possible collection point configuration. 

Three constructive heuristics were defined by prioritizing one of the three objectives of the GAP location problem:
\begin{itemize}
\item \textit{PageRank-Vol} selects the configuration that collects the maximum volume of waste from the nearby generators. If more than one GAP configuration collect the same maximum of waste, the one with the cheapest installation cost is selected;
\item \textit{PageRank-Dist} considers the configurations that have a capacity larger than zero. Then, from these configurations it selects the one that provokes that generators walk the minimal distance (for this it also considers if there is available installed capacity in the nearby locations). If more than one configuration have the same minimum distance, the one that collects the maximum volume of waste is selected.
\item \textit{PageRank-Cost} evaluates the solutions that collects at least all the generated waste by the nearest generator, then it considers the one with the cheapest installation costs. If several configurations have the same minimum cost, the one that collects the maximum volume of waste is selected.
\end{itemize}

\blue{
In addition, a multiobjective version of the PageRank heuristic was developed, accounting for the linear aggregation of the problem objectives. The linear aggregation approach is usually outperformed by Pareto-based methods for multiobjective optimization, but it is a common approach in the related literature mainly because it is efficient and suitable for optimization problems with a convex Pareto front~\cite{Coello2002}.
Instead of taking local decisions by optimizing a single objective function, the multiobjective PageRank (MO-PageRank) considers the linear function $\alpha \times f1 + \beta \times f2 + \gamma \times f3$, being $f1$, $f2$, and $f3$ the objective functions proposed in the mathematical formulation (Eq.~\ref{eq:1}--\ref{eq:3}) and the weights $\{\alpha, \beta, \gamma \} \in \{0.0, 0.1, ..., 0.9, 1.0\}$; $\alpha + \beta + \gamma = 1$. Values of functions $f1$, $f2$, and $f3$ were properly normalized according to the maximum values for each function: i) the maximum cost corresponds to locate a the most expensive bin in all GAPs; ii) the maximum distance is the number of clients $\times$ 300 (the threshold for distance); and the maximum volume corresponds to collecting all the waste generated in the scenario.
}

\subsection{Multiobjective evolutionary algorithms for GAPs location}
\label{subsec:moea}

\paragraph{Evolutionary Algorithms (EAs).} EAs are stochastic soft computing methods that emulate the 
natural evolution to solve optimization, search, and other problems~\cite{BFM97}.
In the last 30 years, EAs have been successfully applied to solve optimization problems underlying many real and complex applications.
Algorithm~\ref{Alg:diagramEA} presents the pseudo-code of an EA.

\begin{algorithm}[!ht]
\caption{Pseudo-code of an Evolutionary Algorithm}
\label{Alg:diagramEA}
\begin{algorithmic}[1]
\State \textit{t} $\leftarrow$ 0 \Comment{generation counter} 
\State {\bf initialize}($P$(0)) \Comment{population initialization}
\While {not stopping\_criterion}
	\State {\bf evaluate}($P$($t$)) \Comment{population evaluation}
	\State parents $\leftarrow$ {\bf selection}($P$(\textit{t}))
	\State offspring $\leftarrow$ {\bf evolutionary operators}(parents)
	\State $P$($t$+1) $\leftarrow$ {\bf replacement}(offspring, $P$($t$))
	\State \textit{t} = \textit{t} + 1
\EndWhile
\State {\bf return} best solution found
\end{algorithmic}
\end{algorithm}

An EA is an iterative technique (each iteration is called a \emph{generation}) that applies stochastic operators on a set of \textit{individuals} (the population \emph{P}), in order to improve their {\em fitness}, a measure related to the objective 
function of the problem.
Every individual in the population is an encoded version of a candidate solution for the problem. The initial population is generated randomly or using a specific heuristic for the problem.
An evaluation function associates a fitness value to every individual.
The search is guided towards higher-quality solutions by a probabilistic selection-of-the-best technique.
Iteratively, solutions are modified by applying \emph{evolutionary operators},
i.e., \textit{recombination} of parts from two individuals or random changes (\textit{mutations}) in their contents, which are applied for building new solutions during the search.
The stopping criterion usually involves a fixed number of generations or execution time, a quality threshold on the best fitness value, or the detection of a stagnation situation. Specific policies are used to select individuals to recombine 
(\emph{selection}) 
and to determine which new individuals are inserted in the population in each new generation 
(\emph{replacement}). 
The EA returns the best solution found in the iterative process, taking into account the fitness function.

\paragraph{Multiobjective Evolutionary Algorithms.} MOEAs~\cite{Coello2002,Deb2001} are evolutionary optimization methods 
conceived to solve problems with two or more conflicting objectives. MOEAs have proven to be efficient in solving difficult real-life optimization problems in many research areas. 
Unlike most traditional methods for multiobjective optimization, MOEAs allow finding a set with several solutions in a single execution, since they work with a population.
%
MOEAs are designed to fulfill two goals at the same time: 
\textit{i}) approximate the Pareto front, using a Pareto-based evolutionary search, and \textit{ii}) maintain diversity, instead of converging to a particular section of the Pareto front, using specific techniques also applied in multi-modal function optimization (e.g., sharing, crowding).

Two state-of-the-art MOEAs are applied in this article: \textit{Non-dominated Sorting Genetic Algorithm, version II} (NSGA-II)~\cite{Deb2001} \blue{and \textit{Strength Pareto Evolutionary Algorithm, version 2} (SPEA2)~\cite{Zitzler2001}. Both NSGA-II and SPEA-2 have been successfully applied in many problems in different application areas.}

NSGA-II is characterized by an evolutionary search using
a non-dominated elitist ordering that diminishes the complexity of the dominance check, a crowding technique for diversity preservation, and a fitness assignment method considering dominance ranks and crowding distance values. NSGA-II has successfully been applied in other smart city problems by our research group, including for waste collection routing~\cite{nesmachnowX2018comparison}, traffic and pollution planning~\cite{Peres2018}, and roadside units location for vehicular networks~\cite{massobrioX2017infrastructure}, among others.
Algorithm~\ref{Alg:diagramNSGA-II} presents the pseudo-code of NSGA-II.

\begin{algorithm}[!ht]
\caption{Pseudo-code of the NSGA-II algorithm}
\label{Alg:diagramNSGA-II}
\begin{algorithmic}[1]
\State $t$ $\leftarrow$ 0 \Comment{generation counter}
\State offspring $\leftarrow$ $\emptyset$ 
	\State {\bf initialize}($P$(0)) \Comment{population initialization}
	\While {not stopping\_criterion}
		\State {\bf evaluate}($P$($t$)) \Comment{population evaluation}
		\State R $\leftarrow$ $P$($t$) $\cup$ offspring
		\State fronts $\leftarrow$ {\bf non-dominated sorting}(R))
		\State $P$($t$+1) $\leftarrow$ $\emptyset$;  $i$ $\leftarrow$ 1
		\While {$\lvert P(t+1)\rvert + \lvert $fronts($i$)$\rvert \leq N$}
		\State {\bf crowding distance}(fronts($i$))
		\State $P$($t$+1) $\leftarrow$ $P$($t$+1) $\cup$ fronts($i$)
		\State$i$ $\leftarrow$ $i$+1
	\EndWhile
	\State {\bf sorting by distance} (fronts(i))
	\State $P$($t$+1) $\leftarrow$ $P$($t$+1) $\cup$ fronts(i)[1:(N - $\lvert$$P$($t$+1)$\rvert$)]
	\State selected $\leftarrow$ \textbf{selection}($P$($t$+1))
	\State offspring $\leftarrow$ \textbf{evolutionary operators}(selected)
	\State $t$  $\leftarrow$  $t$ + 1
\EndWhile
\State {\bf return} computed Pareto front
\end{algorithmic}
\end{algorithm}

\blue{
SPEA2 is a popular state-of-the-art MOEA. It has been successfully applied in many problems in diverse application areas. 
One of the main distinctive features of SPEA2 is the fitness calculation, which is based on both Pareto dominance and diversity: the algorithm defines the \textit{strength} concept to evaluate how many solutions dominate (and are dominated by) each candidate solution, and a density estimation is also considered for fitness assignment. Elitism is also applied, by using an elite population to store the non-dominated individuals found during the search. 

SPEA2 was designed to improve over the main drawbacks of the original SPEA algorithm. The main features of SPEA2 include: \textit{i}) an improved fitness assignment scheme, taking into account for each individual how many individuals it dominates and it is dominated by; \textit{ii}) a nearest neighbor density estimation technique which allows a more precise guidance of the search process; and \textit{iii}) an improved archive truncation method that guarantees the preservation of boundary solutions in the elite population. 

Algorithm~\ref{Alg:SPEA2} presents a schema of SPEA2 working on a population $P$ (size $N$). The elite population is elitePop, having a size of eliteSize. When the elite population is full, a pruning method is applied to remove the most similar individuals to assure that the size of the elite population is always eliteSize.
}

\begin{algorithm}[!ht]
\blue{
\caption{\blue{Schema of the SPEA2 algorithm.}}
\label{Alg:SPEA2}
\begin{algorithmic} [1]
\State  $t$ $\leftarrow$ 0; 	
\State elitePop $\leftarrow$ $\emptyset$
\State  {\bf initialize}($P$(0))
\While {not stopcriterion}
	\State  {\bf evaluate}($P$($t$))
	\State  R $\leftarrow$ $P$($t$) $\cup$ elitePop
	\For {$s_i \in$ R}
		\State $si_{raw}$ $\leftarrow$ computeRawFitness($s_i$,R)
		\State $si_{density}$ $\leftarrow$ computeDensity($s_i$,R)
		\State $si_{fitness}$ $\leftarrow$  $si_{raw} + si_{density}$
	\EndFor
	\State elitePop $\leftarrow$ nonDominated(R)
	\If {size(elitePoP) $>$ eliteSize} 
		\State elitePop $\leftarrow$ removeMostSimilar(elitePop)
	\EndIf
	\State  selected $\leftarrow$ \textbf{selection}(R)
	\State  offspring $\leftarrow$ \textbf{variation operators}(selected)
	\State  $t$  $\leftarrow$  $t$ + 1
\EndWhile
\State {\bf return} computed Pareto front
\end{algorithmic}
}
\end{algorithm}

The proposed NSGA-II \blue{and SPEA-2} for the GAPs location problem include the following features:
\begin{itemize}
\item \textit{Solution encoding}.
Solutions are encoded as a vector of integers in the range [0,$Z$-1]. Each position in the vector represent a possible GAP location (i.e., indexed by $i_1$,...,$i_M$), and the 
integer value on index $i_k$ represents one of the $Z$ possible configurations (i.e., number of bins for each bin type). Configurations are defined taking into account the input data and the space for locating bins in each GAP, according to the problem model and constraints for a given scenario. The special value `0' is used to represent the situation where no bin is installed in a given GAP location (i.e., configuration 0).
Fig.~\ref{Fig:encoding} presents an example of solution encoding for a sample scenario with four 
GAP locations ($i_1, ..., i_4$), two types of bins ($j_1$ and $j_2$), and $Z$ configurations.

\begin{figure}[!h]
\setlength{\abovecaptionskip}{2pt}
    \centering
        \includegraphics[width=0.9\textwidth]{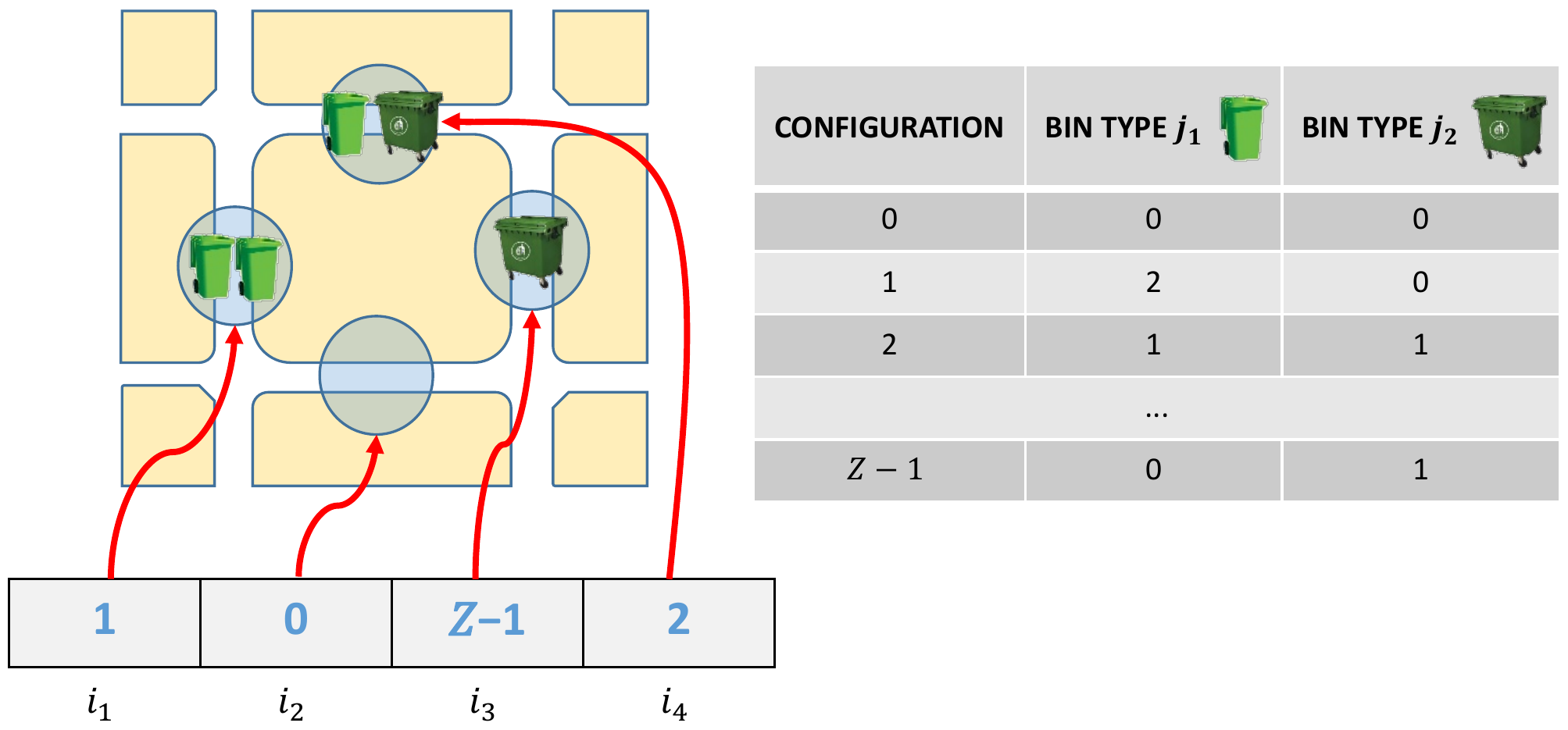}
\caption{Example of solution encoding of an scenario with the visualization of four possible GAP locations and two types of bins.}
\label{Fig:encoding}
\end{figure}

\item \textit{Initialization}. The population is initialized by applying a random procedure that selects a configuration for each GAP, according to a uniform distribution over the $Z$ configurations defined for the problem scenario.

\item \textit{Selection, replacement, and fitness assignment}. NSGA-II applies the ($\mu$+$\lambda$) evolution model. Tournament selection is applied, with tournament size of two individuals. The tournament criteria is based on dominance, and if the two compared individuals are non-dominated, the selection is made based on crowding distance. Fitness assignment is performed considering Pareto dominance rank and crowding distance values.

\item \textit{Evolutionary operators}.
The recombination operator applied is the standard two points crossover (2PX) applied over two selected individuals with probability $p_C$. This operator works as follows: given two parents (individuals), 2PX defines two cutting points by randomly sampling (uniform distribution) two integers in the range [0,$Z$-1] and exchanges between parents the information between the selected cutting points to create two offspring.
The mutation operator is based on randomly modifying specific attributes (i.e., configurations) in a given individual. Elements in a solution encoding are replaced by an integer value, uniformly selected in the range [0,$Z$-1], with probability $p_M$.

\item \textit{Solution feasibility}. \blue{Unlike in the previous two-objective version of the problem~\cite{rossitX2018municipal}, where a constraint stated that all users must have an assigned GAP, in the multiobjective version of the problem all solutions are considered as feasible because of considering distance, waste collected, and cost as explicit objectives in the problem formulation. The only constraint is Eq.~\ref{eq:5} in the problem formulation (available space for installing bins), and all possible bins configurations for GAPs were defined to fulfill the space constraint (see Table~\ref{tab:configurations}). Thus, all solutions generated by the application of the evolutionary operators are feasible.}
\end{itemize}

\section{Experimental evaluation}
\label{sec:exp}

This section reports the experimental analysis of the proposed soft computing methods to solve the real instances of the GAP location problem.

\subsection{Problem instances}

Problem instances were generated by considering data from two real cities: Montevideo, Uruguay and Bah\'ia Blanca, Argentina. 

The MSW system is operated differently in each city. On the one hand, Montevideo has already a community bins based system. The installment of bins started in 2005 with some initial neighborhoods and nowadays reaches the entire urban area of the city. Therefore, the solutions computed by the proposed soft computing methods for these scenarios can be compared with the present configuration of GAPs used by the authorities, in order to suggest ways to improve the distribution of the current collection network. On the other hand, Bah\'ia Blanca still has a door-to-door system where the collection vehicle has to visit every dwelling. However, the City Hall of Bah\'ia Blanca is considering to implement a community bins system that contribute to reduce the logistics cost, which are remarkably high in Argentina~\cite{brozX2018argentinian}. Thus, the solutions proposed for these scenarios can be used as a starting point to migrate from a door-to-door basis to a community bins system. 

The experimental evaluation was performed over four different real urban areas: two from Montevideo (Trouville and Villa Espa\~nola neighborhoods) and two from Bah\'ia Blanca (La Falda and Villa Mitre neighborhoods). 

\begin{figure}[!h]
\setlength{\abovecaptionskip}{6pt}
    \centering
    \begin{subfigure}[b]{0.495\textwidth}
        \includegraphics[width=\textwidth]{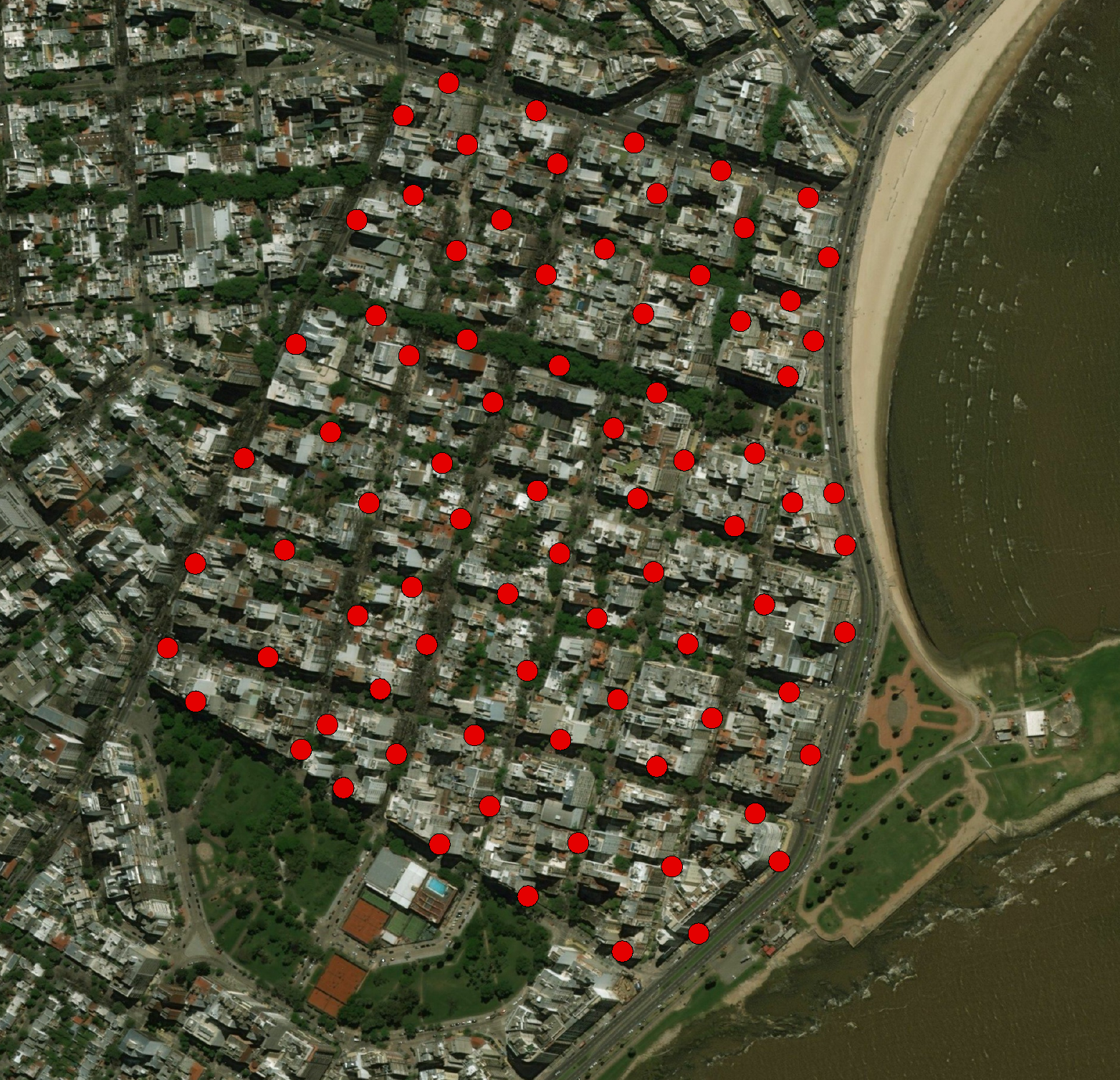}
        \caption{Trouville, Montevideo}
        \label{fig:3D_Trouville}
   \end{subfigure}
    \begin{subfigure}[b]{0.484\textwidth}
        \includegraphics[width=\textwidth]{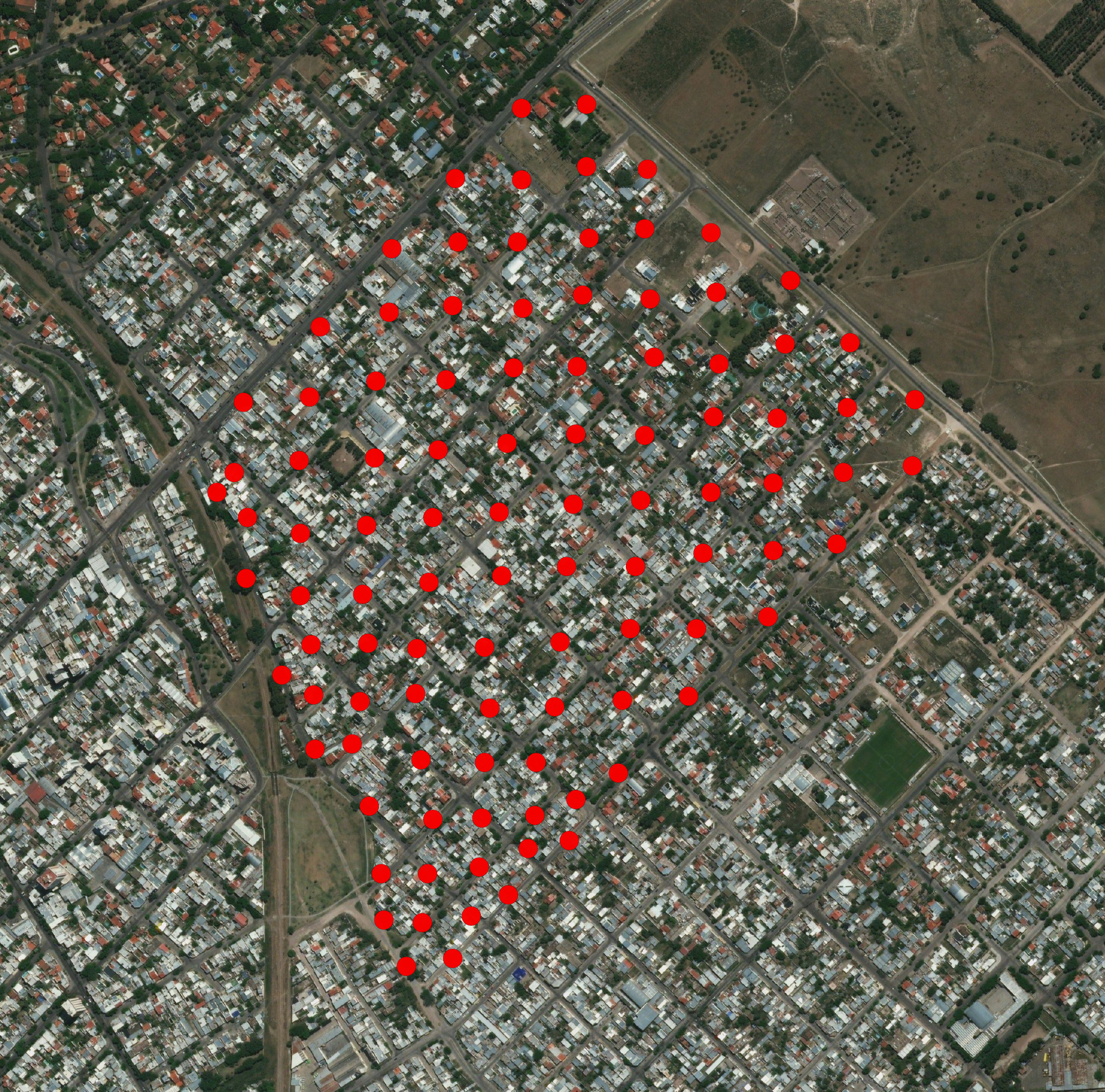}
        \caption{La Falda, Bah\'ia Blanca}
        \label{fig:3D_VEsp}
    \end{subfigure}
\caption{Two of the urban areas studied in this article: Trouville, in Montevideo and La Falda, in Bah\'ia Blanca. Source of the maps: Bing Maps.}
\label{Fig:zones}
\end{figure}

Fig.~\ref{Fig:zones} presents two of the studied areas: Trouville, in Montevideo (including 82 generators) and La Falda, in Bah\'ia Blanca (including 99 generators). For each problem instance, three different scenarios are considered, according to variations of the waste generation rate along the year: a \textit{normal demand} scenario, with the average generation rate estimated by authorities of both cities~\cite{planmaestro2012,plapiqui2013}, a \textit{high demand} scenario, and a \textit{low demand} scenario, with generation rates 20\% larger and smaller than the {normal demand} scenario, respectively. This percentage variation is in line with surveys carried out to practitioners about what may occur along different periods of the year~\cite{CEMPRE}. 

The maximum distance between any generator in $P$ and its assigned GAP is $D = 300$m, in line with suggestions for the maximum distance for accessing to public services. Three bin types ($j_1$, $j_2$, and $j_3$) were considered based on real information about bins that are used by the local government in Montevideo~\cite{MML} (Bah\'ia Blanca does not use community bins nowadays). The values of parameters $c_j$, $C_j$, and $e_j$ for each type are: 1000 monetary units (m.u.), 1 $m^{3}$ and 1 $m^{2}$ for bin type $j_1$, 2000 m.u., 2 $m^{3}$ and 2 $m^{2}$ for bin type $j_2$, and 3000 m.u., 3 $m^{3}$ and 3 $m^{2}$ for bin type $j_3$. 
Details of the possible configurations for each GAP, including the required space for installation, total cost, and capacity, are presented in Table~\ref{tab:configurations}.
\vspace{3mm}

\begin{table}[!h]
	\setlength{\tabcolsep}{10.5pt}
	\setlength{\abovecaptionskip}{3pt}
    \renewcommand{\arraystretch}{0.9}
	\small
	\centering    
    \caption{Feasible configurations according to the problem definition and GAP constraints ($S_i$ = 5 for all collection points).}
    \label{tab:configurations}
    \begin{tabular}{rrrrrrr}
    \toprule
\multicolumn{1}{c}{\textit{config.}} & \multicolumn{3}{c}{\textit{number of bins}} & \multicolumn{1}{c}{\textit{required}} & \multicolumn{1}{c}{\textit{installation}} & \multicolumn{1}{c}{\textit{maximum}} \\
\cline{2-4}\\[-9pt]
\multicolumn{1}{c}{\textit{id}} & {$j_1$} & {$j_2$} & {$j_3$} & \multicolumn{1}{c}{\textit{space} ($m^2$)} & \multicolumn{1}{c}{\textit{cost (m.u.)}} & \multicolumn{1}{c}{\textit{capacity} ($m^3$)} \\
\midrule
0  &  0  &  0  &  0  &  0 &  0  &  0  \\
1  &  1  &  0  &  0  &  1 &  1000  &  1  \\
2  &  2  &  0  &  0  &  2 &  2000  &  2  \\
3  &  3  &  0  &  0  &  3 &  3000  &  3  \\
4  &  4  &  0  &  0  &  4 &  4000  &  4  \\
5  &  5  &  0  &  0  &  5 &  5000  &  5  \\
6  &  1  &  1  &  0  &  3 &  3000  &  3  \\
7  &  1  &  2  &  0  &  5 &  5000  &  5  \\
8  &  1  &  0  &  1  &  4 &  4000  &  4  \\
9  &  0  &  1  &  0  &  2 &  2000  &  2  \\
10  &  0  &  1  &  1  &  5 &  5000  &  5  \\
11  &  0  &  0  &  1  &  3 &  3000  &  3  \\
\bottomrule
\end{tabular}
\end{table}

\vspace{-4mm}

\subsection{Methodology}

\paragraph{Development and execution platform.}
The proposed methods were implemented in Python, by using the Distributed Evolutionary Algorithms in Python (deap) framework~\cite{fortin2012deap}.
The experimental evaluation was performed on a Dell PowerEdge M620 (Intel Xeon E5-2680 processor at 2.50GHz, 24 cores and 32 GB RAM) from Cluster FING, 
Universidad de la Rep\'ublica, Uruguay~\cite{nesmachnow2010cluster}.

\paragraph{Metrics.}
Results computed by the proposed MOEAs are evaluated considering two standard metrics for multiobjective optimization: \textit{relative hypervolume} (RHV) \blue{ and \textit{spread}}.

RHV is defined as the ratio between the volumes (in the objective functions space) covered by the computed Pareto front and the true Pareto front of the problem. The ideal value for RHV is 1. 
\blue{ 
It is a \textit{combined} metric that evaluates both the numerical accuracy (proximity to the Pareto front) and the distribution of the computed Pareto front. 

Spread is a full \emph{diversity} metric that measures the distribution of the computed non-dominated solutions, evaluating the capability of correctly sampling the Pareto front, Unlike spacing, the spread metric includes the information about the extreme points of the true Pareto front in order to compute a more precise value of the distribution, as defined by Eq.~\ref{Eq:Spr}~\cite{Deb2001}. Smaller values of the spread metric mean a better distribution of non-dominated solutions in the calculated Pareto front, and the metric takes a value zero for an ideal equally-spaced distribution.
\begin{equation}
\label{Eq:Spr}
spread = \frac{\sum\limits^{k}_{h=1}{d_h}^e + \sum\limits^{ND}_{i=1} \left( \bar{d}-d_i\right)^2}{\sum\limits^{k}_{h=1}{d_h}^e + ND \times \bar{d}}
\end{equation}
}

In the analysis, the true Pareto front---unknown for the problem instances studied---is approximated by all non-dominated solutions found for each instance in each execution \blue{of each multiobjective algorithm studied}.

\subsection{Numerical results}
\label{subsec:results}

This section reports and analyzes the results obtained by the proposed soft computing methods to solve the GAPs location problem.

\subsubsection{Parameters calibration}
\blue{Parameters calibration is a relevant issue for any stochastic optimization \sloppy method.}
A set of parametric setting experiments were performed to determine the best parameter values for the proposed MOEAs. The parameter setting analysis were made over three instances (different from validation scenarios) to avoid bias in the results. 

{
Performing an exhaustive parameter calibration (e.g., by using automatic methods such as irace~\cite{Lopez2016} and/or SMAC~\cite{Hutter2011} is beyond the scope of this article, so we followed the standard approach of analyzing a set of representative values for the main parameters of the proposed MOEAs (population size, stopping criterion, and probabilities of application of the evolutionary operators) and study the Cartesian product of them to determine the best values.
}

The population size ($\#p$) and the maximum number of generations ($\#g$) were calibrated in preliminary experiments. The analysis confirmed that using $\#p$ = 100 and $\#g$ = 1000 provided a good exploration pattern, which allowed computing the best results. {The same values of $\#p$ and $\#g$ were used for NSGA-II and SPEA-2, to allow a fair results comparison. In SPEA-2, the size of the elite population was set to 20 individuals, following rules-of-thumb from the related literature~\cite{Zitzler2001}.}

For the crossover probability ($p_{C}$) and the mutation probability ($p_{M}$), three different candidate values were defined and all combinations of $p_{C}$ and $p_{M}$ were studied on 30 independent executions performed for {the proposed MOEAs}. Candidate values were $p_{C} \in \{0.5, 0.7, 0.9\}$ and $p_{M} \in \{0.01, 0.05, 0.1\}$. The result distributions obtained using each configuration were analyzed by applying the non-parametric Friedman rank statistical test to determine the configuration that allowed computing the best results.

For NSGA-II, the three best configurations were (0.9, 0.01), (0.7, 0.01), and (0.5, 0.01), with $\chi^2$ = 231.9 and $p$-value $<$ 10$^{-10}$. {For SPEA-2, the three best configurations were (0.9, 0.01), (0.7, 0.01), and (0.7, 0.05), with $\chi^2$ = 202.4 and $p$-value $<$ 10$^{-10}$. A post-hoc analysis applying the Wilcoxon rank test for pairwise comparisons reported that $p_{C}$ = 0.9, $p_{M}$ = 0.01 is the best configuration for both MOEAs, with statistical confidence of 0.97.}

\subsubsection{Multiobjective optimization analysis}

\begin{figure}[!h]
\setlength{\abovecaptionskip}{6pt}
    \centering
    \begin{subfigure}[b]{0.495\textwidth}
        \includegraphics[width=\textwidth, trim={1.2cm 0.8cm 1.2cm 1.2cm},clip]{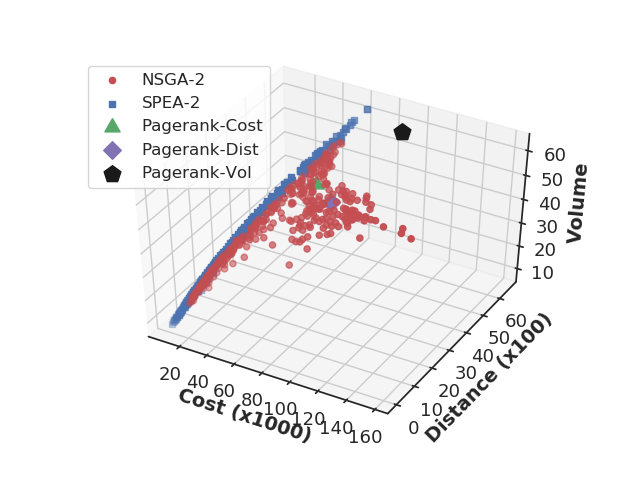}
        \caption{Trouville}
        \label{fig:3D_Trouville}
   \end{subfigure}
    \begin{subfigure}[b]{0.495\textwidth}
        \includegraphics[width=\textwidth, trim={1.2cm 0.8cm 1.2cm 1.2cm},clip]{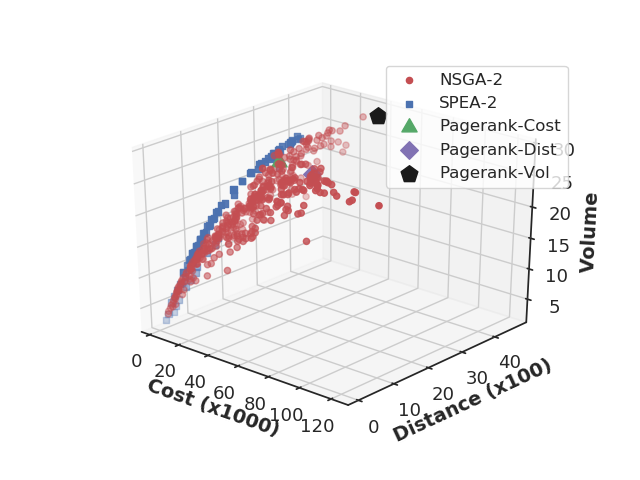}
        \caption{Villa Espa\~nola}
        \label{fig:3D_VEsp}
    \end{subfigure}
    \begin{subfigure}[b]{0.495\textwidth}
        \includegraphics[width=\textwidth, trim={1.2cm 0.8cm 1.0cm 1.2cm},clip]{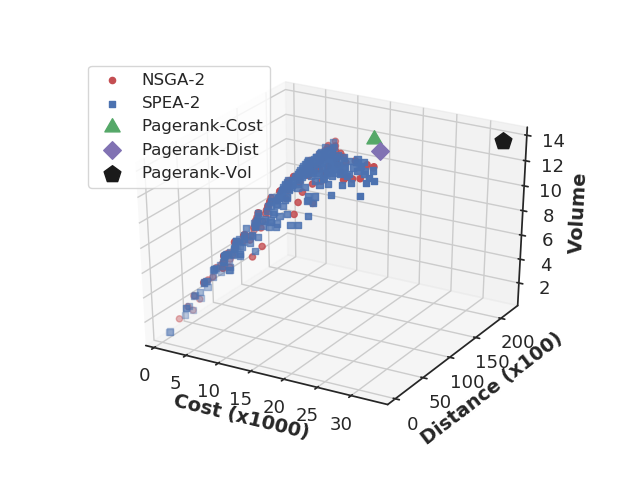}
        \caption{La Falda}
        \label{fig:3D_BBCA2}
   \end{subfigure}
    \begin{subfigure}[b]{0.495\textwidth}
        \includegraphics[width=\textwidth, trim={1.2cm 0.8cm 1.0cm 1.2cm},clip]{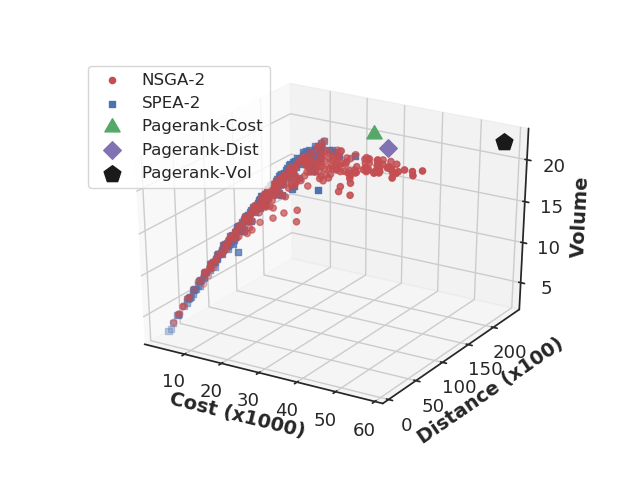}
        \caption{Villa Mitre}
        \label{fig:3D_BBCA3}
    \end{subfigure}
\caption{{Sample Pareto fronts computed by NSGA-II and SPEA-2 and PageRank solutions for the four scenarios with normal waste generation.}}
\label{Fig:Pareto3D}
\end{figure}

\paragraph{{Sample Pareto fronts.}} 
Fig.~\ref{Fig:Pareto3D} presents sample Pareto fronts (3D view) computed by {the proposed MOEAs} for the four scenarios with normal waste generation. These results are representative of the results computed for the other waste generation rates. The results computed {by the PageRank heuristics} are also reported for comparative purposes.

Results in Fig.~\ref{Fig:Pareto3D} indicate that {both NSGA-II and SPEA are} able to accurately sample the set of trade-off solutions for the problem instances studied.
PageRank-Dist and PageRank-Cost computed accurate solutions regarding both distance and cost objectives, while PageRank-Vol computed solutions with significantly worst cost and QoS. The proposed {MOEAs take} advantage of the evolutionary search to compute better solutions than PageRank regarding distance and cost, while also sampling properly the Pareto front of the problem. {The graphics also show that NSGA-II is able to compute more non-dominated points than SPEA-2 and also to provide a better coverage of the Pareto front for three out of four scenarios. These visual results are consistent with the analysis of multiobjective optimization metrics presented in the next paragraph.}

\paragraph{{Multiobjective optimization metrics comparison.}}

{The following methodology was applied to analyze the hypervolume and spread results distributions of the proposed multiobjective methods. First, the Kolmogorov-Smirnov statistical test was applied to check normality of the results distribution. As the results of the statistical tests did not confirm normality for any metric or algorithm, the median value was used as estimator and the results distributions were further studied using the non-parametric Wilcoxon test to analyze the statistical significance of the differences on the means of the hypervolume and spread values, considering a p-value of 0.05. Finally, in those cases where no conclusive difference was found (with confidence level of 0.95) the interquartile range was studied as a metric of deviation (for non-uniform distributions) to compare the means of distributions.}


Table~\ref{tab:hvn_median} reports the hypervolume results obtained by {NSGA-II, SPEA-2, and MO-PageRark (MO-PR in the table)} for each of the {twelve} problem instances studied. The median value is used as estimator, because results do not follow a normal distribution. {The best median values are marked in bold font. The asterisk ($\star$) means that there is statistical significance according to the Wilcoxon test (confidence level of 0.95) and the circle ($\circ$) means that the difference between medians is larger than the interquartile range.
}

\begin{table*}[!h]
	\setlength{\tabcolsep}{5pt}
	\centering
	\small
	\caption{{Hypervolume results for the studied problem instances}}
	\vspace{-0.2cm}
    \label{tab:hvn_median}
    {\color{black}
	\begin{tabular}{@{}lcrrrrrrrrr@{}}
	    \toprule
	\multirow{2}{*}{\textit{scenario}} & \multirow{2}{*}{\textit{waste}} & \multicolumn{3}{c}{NSGA-2} &&  \multicolumn{3}{c}{SPEA-2} & \multirow{2}{*}{{MO-PR}}\\
	\cline{3-5} \cline{7-9}\\[-9pt]
	& 
	& \textit{min.} & \textit{med.} & \textit{max.} & & \textit{min.} & \textit{med.} & \textit{max.}  \\
\midrule
\multirow{3}{*}{Trouville} & low & 0.973 & 0.983  & 0.993  && 0.979 & $\star$\textbf{0.985} & 0.993 & 0.028  \\ 
 & normal & 0.965 & 0.977  & 0.986  && 0.965 & $\circ$\textbf{0.979} & 0.984 & 0.096  \\ 
 & high & 0.968 & $\circ$\textbf{0.985}  & 0.988  && 0.975 & 0.983  & 0.987 & 0.108  \\ 
 \midrule
\multirow{3}{*}{V.~Espa\~nola} & low & 0.990 & $\star$\textbf{0.996}  & 0.999  && 0.991 & 0.994  & 0.997 & 0.081  \\ 
 & normal & 0.988 & \textbf{0.994}  & 0.999  && 0.991 & 0.993  & 0.997 & 0.056  \\ 
 & high & 0.977 & 0.992  & 0.997  && 0.990 & \textbf{0.993}  & 0.996 & 0.082  \\ 
\midrule
\multirow{3}{*}{La Falda} & low & 0.934 & $\star$\textbf{0.981}  & 0.995  && 0.943 & 0.977  & 0.987 & 0.036  \\ 
 & normal & 0.938 & $\star$\textbf{0.983}  & 0.995  && 0.943 & 0.979  & 0.991 & 0.037  \\ 
 & high & 0.969 & $\star$\textbf{0.985}  & 0.993  && 0.949 & 0.981  & 0.992 & 0.044  \\ 
\midrule
\multirow{3}{*}{V.~Mitre}& low & 0.955 & 0.976  & 0.983  && 0.953 & {0.976}  & 0.996 & 0.038  \\ 
 & normal & 0.943 & \textbf{0.976}  & 0.989  && 0.962 & 0.975  & 0.989 & 0.036  \\ 
 & high & 0.955 & $\star$\textbf{0.982}  & 0.986  && 0.947 & 0.974  & 0.986 & 0.037  \\ 
		\bottomrule
	\end{tabular}
	}
	\vspace{0.5cm}
\end{table*}

Results in Table~\ref{tab:hvn_median} indicate that {both proposed MOEAs were} able to consistently compute accurate solutions for all problem instances studied. Hypervolume results {of both MOEAs} were over 0.93 for all cases. The best hypervolume results computed were 0.98 or superior, demonstrating the robustness of the proposed evolutionary search. 
%
{On the other hand, MO-PageRank was not able to compute good hypervolume values, suggesting that the linear aggregation approach is not useful to solve the problem.}

{Regarding the comparison between NSGA-II and SPEA-2, results show that both algorithms computed similar values of the hypervolume metric. However, NSGA-II outperformed SPEA-2 in eight out of twelve problem instances, in five of them with statistical significance according to the Wilcoxon test.
Results also confirmed that the variations of waste generation rates within each scenario did not affect significantly the hypervolume results.
}

{Table~\ref{tab:spread} reports the spread values for the proposed multiobjective algorihtms. The median value is used as estimator, because results do not follow a normal distribution. Again, the best median values are marked in bold font and the asterisk ($\star$) means that there is statistical significance according to the Wilcoxon test (confidence level of 0.95).}

\begin{table*}[!h]
	\setlength{\tabcolsep}{5pt}
	\centering
	\caption{{Spread results for the studied problem instances}}
	\vspace{-0.2cm}
    \label{tab:spread}
	\small
	{\color{black}
	\begin{tabular}{@{}lcrrrrrrrrr@{}}
	    \toprule
	\multirow{2}{*}{\textit{scenario}} & \multirow{2}{*}{\textit{waste}} & \multicolumn{3}{c}{NSGA-2} &&  \multicolumn{3}{c}{SPEA-2} & \multirow{2}{*}{{MO-PR}}\\
	\cline{3-5} \cline{7-9}\\[-9pt]
	& 
	& \textit{min.} & \textit{med.} & \textit{max.} & & \textit{min.} & \textit{med.} & \textit{max.}  \\
\midrule
\multirow{3}{*}{Trouville} & low & 0.722 & 0.938  & 0.996  && 0.652 & $\star$\textbf{0.854}  & 0.999 & 0.500  \\ 
 & normal & 0.515 & $\star$\textbf{0.710}  & 0.992  && 0.523 & 0.823  & 0.998 & 0.333  \\ 
 & high & 0.372 & $\star$\textbf{0.643}  & 0.971  && 0.495 & 0.771  & 0.987 & 0.750  \\ 
 \midrule
\multirow{3}{*}{V.~Espa\~nola} & low & 0.833 & 0.928  & 0.999  && 0.680 & \textbf{0.876}  & 0.999 & 0.600  \\ 
 & normal & 0.693 & $\star$\textbf{0.878}  & 0.983  && 0.722 & 0.922  & 0.999 & 0.500 \\ 
 & high & 0.597 & \textbf{0.828}  & 0.979  && 0.605 & 0.864  & 0.986 & 0.750  \\ 
\midrule
\multirow{3}{*}{La Falda} & low & 0.404 & 0.572  & 0.754  && 0.322 & $\star$\textbf{0.464}  & 0.614 & 0.429 \\ 
 & normal & 0.326 & \textbf{0.564 } & 0.688  && 0.331 & 0.586  & 0.974 & 0.429\\ 
 & high & 0.377 & 0.582  & 0.731  && 0.318 & $\star$\textbf{0.388}  & 0.775 & 0.429 \\ 
\midrule
\multirow{3}{*}{V.~Mitre}& low & 0.479 & 0.625  & 0.705  && 0.406 & \textbf{0.584}  & 0.840 & 0.795 \\ 
 & normal & 0.464 & 0.613  & 0.763  && 0.400 & \textbf{0.572}  & 0.767 & 0.873 \\ 
 & high & 0.548 & $\star$\textbf{0.646}  & 0.744  && 0.561 & 0.719  & 0.976 & 0.729  \\ 
		\bottomrule
	\end{tabular}
	}
\end{table*}

\newpage

{Results in Table~\ref{tab:spread} show that both MOEAs have similar values of the spread metric, suggesting a correct distribution of non-dominated points in the computed Pareto fronts. MO-PR has better values of the spread metric, but it is not considered as the best method because of its poor results (as reported for the hypervolume analysis) and for the significantly few non-dominated points in the computed Pareto front (less than 10 points in average).
NSGA-II outperformed SPEA-2 in six instances (four with statistical significance) and SPEA-2 had the best spread values in other six instances (three with statistical significance).

Taking into account the reported results, and considering that both studied MOEAs had a similar performance in terms of solution quality and diversity, with NSGA-II computing better hypervolume and spread values in most cases, NSGA-II was selected for the comparative analysis of each objective funtion, presented in the following subsection.}

\subsubsection{Comparative analysis}
Table~\ref{tab:improv} reports the improvements of the {NSGA-II} results over the {single- and multi-objective} PageRank heuristics for the {twelve} problem instances {on the four scenarios} studied.
The reported values accounts for the average and best improvements in each one of the three problem objectives (distance, cost, and volume) over each PageRank solution, computed over those solutions in the Pareto front {of NSGA-II} that dominate the corresponding PageRank solution in distance and cost objectives, and has up to 10\% difference on the volume of the collected waste. {In the case of the MO-PageRank, the solution compared with NSGA-II is the one which is the closest to the ideal vector (computed from the Pareto front of the problem~\cite{Deb2001}.}
Given that most of the waste is collected in the solutions computed by all PageRank heuristics and {NSGA-II}, the analysis is focused on the benefits for both citizens (i.e., QoS, given by the average distance they must walk to dispose the waste) and the city administration (evaluating the cost of implementing a certain GAP planning).

\begin{table}[!h]
\setlength{\tabcolsep}{3pt}
\renewcommand{\arraystretch}{0.9}
\small
\centering    
\caption{Improvements of the {NSGA-II} solutions over the PageRank solutions.}
\label{tab:improv}
\begin{tabular}{cclrrrrrr}
\toprule
\multirow{2}{*}{\textit{scenario}} & \multicolumn{1}{c}{\textit{waste}} & \multicolumn{1}{c}{\multirow{2}{*}{\textit{baseline}}} & \multicolumn{3}{c}{\textit{average improvement}} & & \multicolumn{2}{c}{\textit{best improvement}}  \\
\cline{4-6}\cline{8-9}
& \multicolumn{1}{c}{\textit{generation}} & & \multicolumn{1}{c}{\textit{distance}} & \multicolumn{1}{c}{\textit{cost}} & \multicolumn{1}{c}{\textit{volume}} & & \multicolumn{1}{c}{\textit{distance}} & \multicolumn{1}{c}{\textit{cost}}\\
\midrule
 \parbox[t]{2mm}{\multirow{12}{*}{\rotatebox[origin=c]{90}{Trouville}}}
& \multirow{2}{*}{low} 
  & PageRank-Cost & 6.0\% & 8.0\% & 8.7\% && 15.6\%  & 13.6\%   \\ 
& \multirow{2}{*}{demand} 
  & PageRank-Dist & 9.9\% & 7.3\% & 8.9\% && 33.5\%  & 14.1\%   \\ 
  & & PageRank-Vol & 44.0\% & 17.7\% & 5.1\% && 79.5\%  & 37.2\%   \\ 
  & & {MO-PageRank} & {48.2\%} & {81.4\%} & {3.1\%} &  & {63.7\%} & {92.1\%}   \\ 
\cline{2-9}\\[-7pt]
& \multirow{2}{*}{normal} 
  & PageRank-Cost & 16.8\% & 9.4\% & 6.7\% && 38.0\%  & 20.0\%   \\ 
& \multirow{2}{*}{demand} 
  & PageRank-Dist & 18.0\% & 9.9\% & 6.6\% && 36.5\%  & 26.6\%   \\ 
  & & PageRank-Vol & 44.8\% & 23.8\% & 4.7\% && 76.3\%  & 44.4\%   \\ 
  & & {MO-PageRank} & {30.0\%} & {83.5\%} & {1.8\%} &  & {51.6\%} & {90.4\%}   \\ 
\cline{2-9}\\[-7pt]
 & \multirow{2}{*}{high}  
  & PageRank-Cost & 18.1\% & 10.4\% & 8.5\% && 33.3\%  & 20.2\%   \\ 
& \multirow{2}{*}{demand} 
  & PageRank-Dist & 14.8\% & 10.5\% & 8.4\% && 22.1\%  & 21.0\%   \\ 
  & & PageRank-Vol & 47.5\% & 18.9\% & 4.0\% && 80.7\%  & 35.5\%   \\ 
  & & {MO-PageRank} & {29.8\%} & {84.4\%} & {4.6\%} &  & {46.1\%} & {92.4\%}   \\ 
\midrule
\parbox[t]{2mm}{\multirow{12}{*}{\rotatebox[origin=c]{90}{Villa Espa\~nola}}}
& \multirow{2}{*}{low} 
  & PageRank-Cost & 9.3\% & 0.0\% & 9.1\% && 9.3\%  & 0.0\%   \\ 
& \multirow{2}{*}{demand} 
  & PageRank-Dist & 16.8\% & 12.1\% & 4.4\% && 36.6\%  & 37.1\%   \\
  & & PageRank-Vol & 23.7\% & 14.0\% & 3.8\% && 59.1\%  & 33.3\%   \\ 
 & & {MO-PageRank} & {60.5\%} & {83.4\%} & {4.0\%} &  & {83.3\%} & {92.4\%}   \\ 
\cline{2-9}\\[-7pt]
& \multirow{2}{*}{normal}
& PageRank-Cost & 2.3\% & 9.2\% & 7.1\% && 4.1\%  & 18.4\%   \\ 
& \multirow{2}{*}{demand} & PageRank-Dist & 19.8\% & 10.8\% & 6.0\% && 40.0\%  & 22.2\%   \\ 
& & PageRank-Vol & 31.8\% & 13.0\% & 4.8\% && 66.0\%  & 25.6\%   \\ 
  & & {MO-PageRank} & {45.0\%} & {84.3\%} & {4.3\%} &  & {75.0\%} & {93.4\%}   \\ 
\cline{2-9}\\[-7pt]
 & \multirow{2}{*}{high} 
   & PageRank-Cost & 10.0\% & 7.4\% & 6.2\% && 11.7\%  & 7.4\%   \\ 
& \multirow{2}{*}{demand} 
  & PageRank-Dist & 16.8\% & 12.4\% & 5.6\% && 31.7\%  & 25.3\%   \\ 
   & & PageRank-Vol & 36.3\% & 19.8\% & 5.1\% && 69.7\%  & 37.0\%   \\ 
  & & {MO-PageRank} & {41.3\%} & {82.7\%} & {5.1\%} &  & {71.7\%} & {92.6\%}   \\ 
 \hline
 \parbox[t]{2mm}{\multirow{12}{*}{\rotatebox[origin=c]{90}{La Falda}}}
& \multirow{2}{*}{low} 
  & PageRank-Cost & 29.4\% & 6.0\% & 3.3\% && 49.5\%  & 27.8\%   \\ 
& \multirow{2}{*}{demand} 
  & PageRank-Dist  & 14.4\% & 3.9\% & 1.6\% && 49.1\%  & 31.6\%\\ 
  & & PageRank-Vol & 15.3\% & 12.1\% & 1.1\% && 61.2\%  & 60.6\%   \\ 
  & & {MO-PageRank} & {81.7\%} & {53.3\%} & {2.7\%} &  & {88.4\%} & {62.0\%}   \\ 
\cline{2-9}\\[-7pt]
& \multirow{2}{*}{normal} 
  & PageRank-Cost & 34.9\% & 6.8\% & 3.2\% && 53.9\%  & 27.3\%   \\
& \multirow{2}{*}{demand} 
  & PageRank-Dist & 16.7\% & 4.3\% & 1.6\% && 52.7\%  & 30.4\%   \\ 
  & & PageRank-Vol & 16.7\% & 10.8\% & 1.1\% && 64.7\%  & 55.6\%   \\ 
  & & {MO-PageRank} & {81.8\%} & {54.8\%} & {2.9\%} &  & {87.3\%} & {64.3\%}   \\ 
\cline{2-9}\\[-7pt]
 & \multirow{2}{*}{high}  
  & PageRank-Cost & 46.5\% & 8.2\% & 3.5\% && 63.3\%  & 26.9\%   \\  
& \multirow{2}{*}{demand} 
  & PageRank-Dist & 22.5\% & 5.0\% & 1.7\% && 62.2\% & 29.6\%   \\ 
  & & PageRank-Vol & 17.8\% & 13.0\% & 1.2\% && 68.1\%  & 56.8\%   \\ 
  & & {MO-PageRank} & {78.9\%} & {59.7\%} & {3.6\%} &  & {86.2\%} & {68.7\%}   \\ 
\midrule
 \parbox[t]{2mm}{\multirow{9}{*}{\rotatebox[origin=c]{90}{Villa Mitre}}}
& \multirow{2}{*}{low} 
  & PageRank-Cost & 35.0\% & 99.9\% & 3.3\% && 52.0\%  & 99.9\%   \\ 
& \multirow{2}{*}{demand} 
  & PageRank-Dist & 15.9\% & 49.5\% & 1.7\% && 49.7\%  & 99.9\%    \\
  & & PageRank-Vol &15.4\% & 33.3\% & 1.1\% && 60.4\%  & 99.9\%  \\ 
   & & {MO-PageRank} & {76.3\%} & {54.3\%} & {2.9\%} &  & {85.6\%} & {64.3\%}   \\
\cline{2-9}\\[-7pt]
& \multirow{2}{*}{normal}
& PageRank-Cost & 28.4\% & 99.9\% & 3.3\% && 50.1\%  & 99.9\%    \\ 
f& \multirow{2}{*}{demand} & PageRank-Dist & 15.4\% & 49.9\% & 1.6\% && 51.8\%  & 99.9\%  \\ 
& & PageRank-Vol & 16.3\% & 33.3\% & 1.1\% && 64.3\%  & 100.0\%    \\
  & & {MO-PageRank} & {74.8\%} & {57.9\%} & {3.8\%} &  & {83.4\%} & {68.9\%}   \\
\cline{2-9}\\[-7pt]
 & \multirow{2}{*}{high} 
   & PageRank-Cost & 32.1\% & 99.9\% & 3.4\% && 54.8\%  & 99.9\%    \\ 
& \multirow{2}{*}{demand} 
  & PageRank-Dist & 15.7\% & 49.9\% & 1.6\% && 54.4\%  & 99.9\%  \\ 
   & & PageRank-Vol & 17.9\% & 33.3\% & 1.1\% && 69.3\%  & 99.9\%   \\
   & & {MO-PageRank} & {77.0\%} & {58.2\%} & {3.3\%} &  & {82.3\%} & {71.8\%}   \\
\bottomrule
\vspace{3mm}
\end{tabular}
\end{table}

\newpage 

The analysis of results in Table~\ref{tab:improv} allows concluding that {NSGA-II} is able to compute solutions that account for significant improvements over the PageRank algorithms. 

In Montevideo scenarios, regarding the distance objective, the average improvement over PageRank-Dist was up to 18\% (Trouville scenario, normal demand) and the best improvement was 40.0\% (Villa Espa\~nola scenario, normal demand). Regarding the cost objective, the average improvement over PageRank-Cost was up to 10.4\% (Trouville scenario, high demand) and the best improvement was 38.0\% (Trouville scenario, normal demand). Overall, the solutions computed by the proposed {NSGA-II} outperformed the PageRank solutions in all but one case (PageRank-Cost, Villa Espa\~nola scenario, low demand). 

In Bah\'ia Blanca scenarios, regarding the distance objective, the average improvement over PageRank-Dist was up to 22.53\% (La Falda scenario, high demand) and the best improvement was 62.24\% (also La Falda scenario, high demand). Regarding the cost objective, the average improvement over PageRank-Cost was up to 99.89\% (Villa Mitre scenario, normal and high demand) and the best improvement was 63.26\% (La Falda scenario, high demand). For all the scenarios in Bah\'ia Blanca, the solutions computed by the proposed {NSGA-II} outperformed the PageRank solutions.

{The comparison with MO-PageRank results shows a clear superiority of NSGA-II, with improvements of up 93.4\% in cost (for Villa Espa\~nola, normal demand scenario) and up to 88.4\% in distance (for La Falda, low demand scenario). These results are explained because MO-PageRank failed to compute accurate solutions when compared with the Pareto front computed by the proposed MOEAs.}

\begin{figure}[!h]
\setlength{\abovecaptionskip}{3pt}
\centering    
\begin{tabular}{cc}
\includegraphics[width=0.5\textwidth]{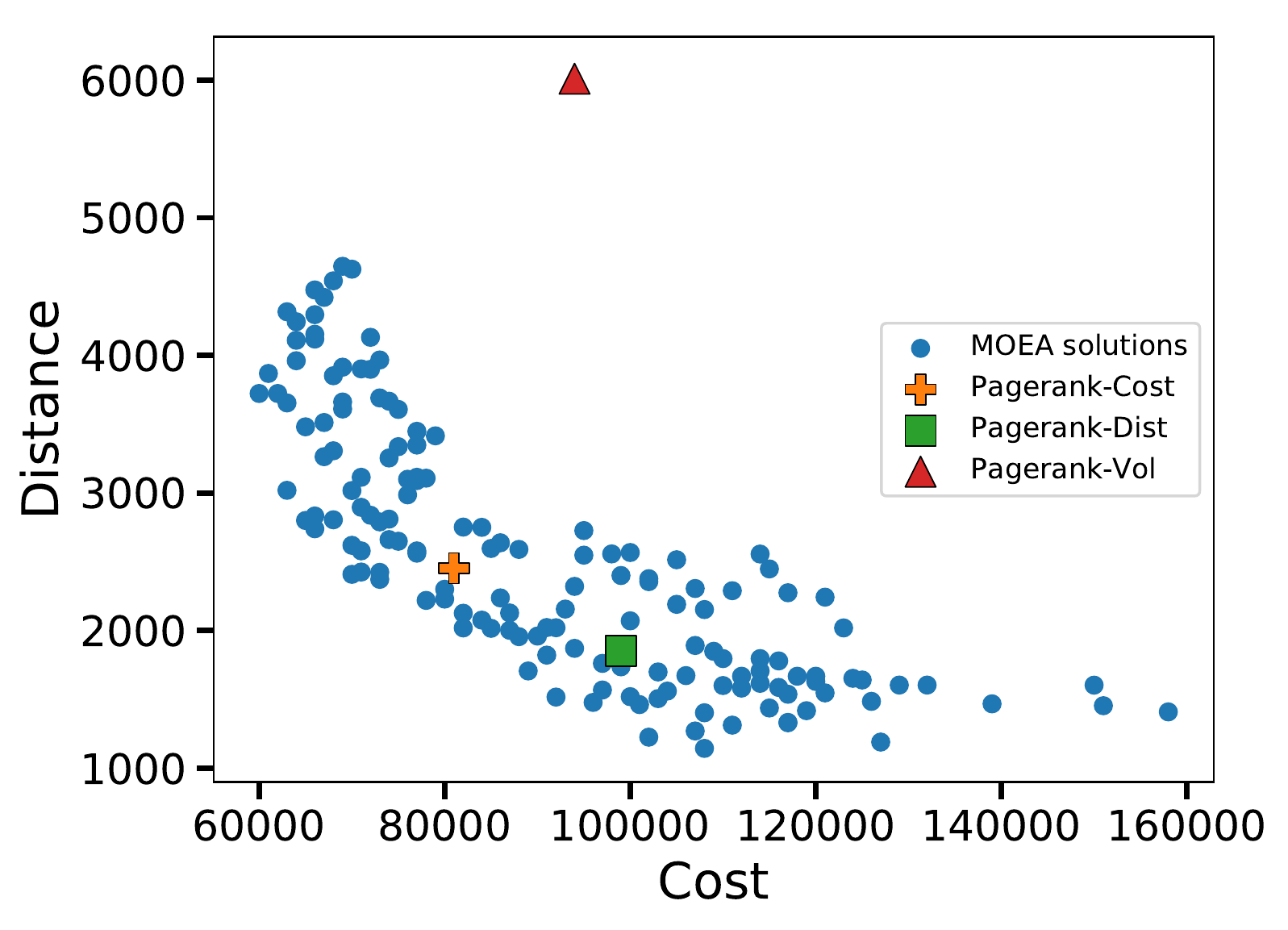} & \includegraphics[width=0.5\textwidth]{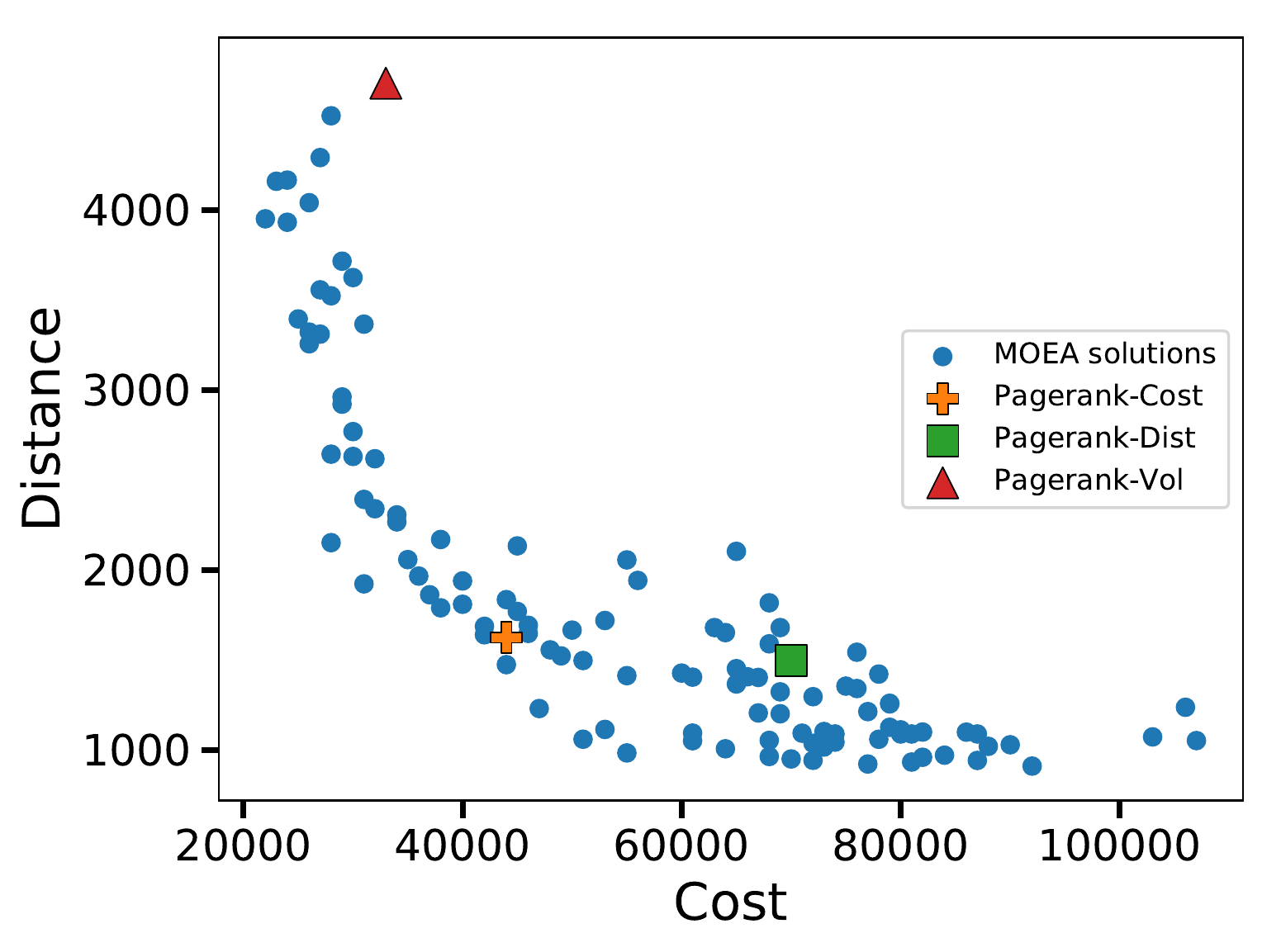}\\[-2pt]
\small{(a1) Trouville, low demand} & \small{(b1) Villa Espa\~nola, low demand}\\[-1pt]
\includegraphics[width=0.515\textwidth]{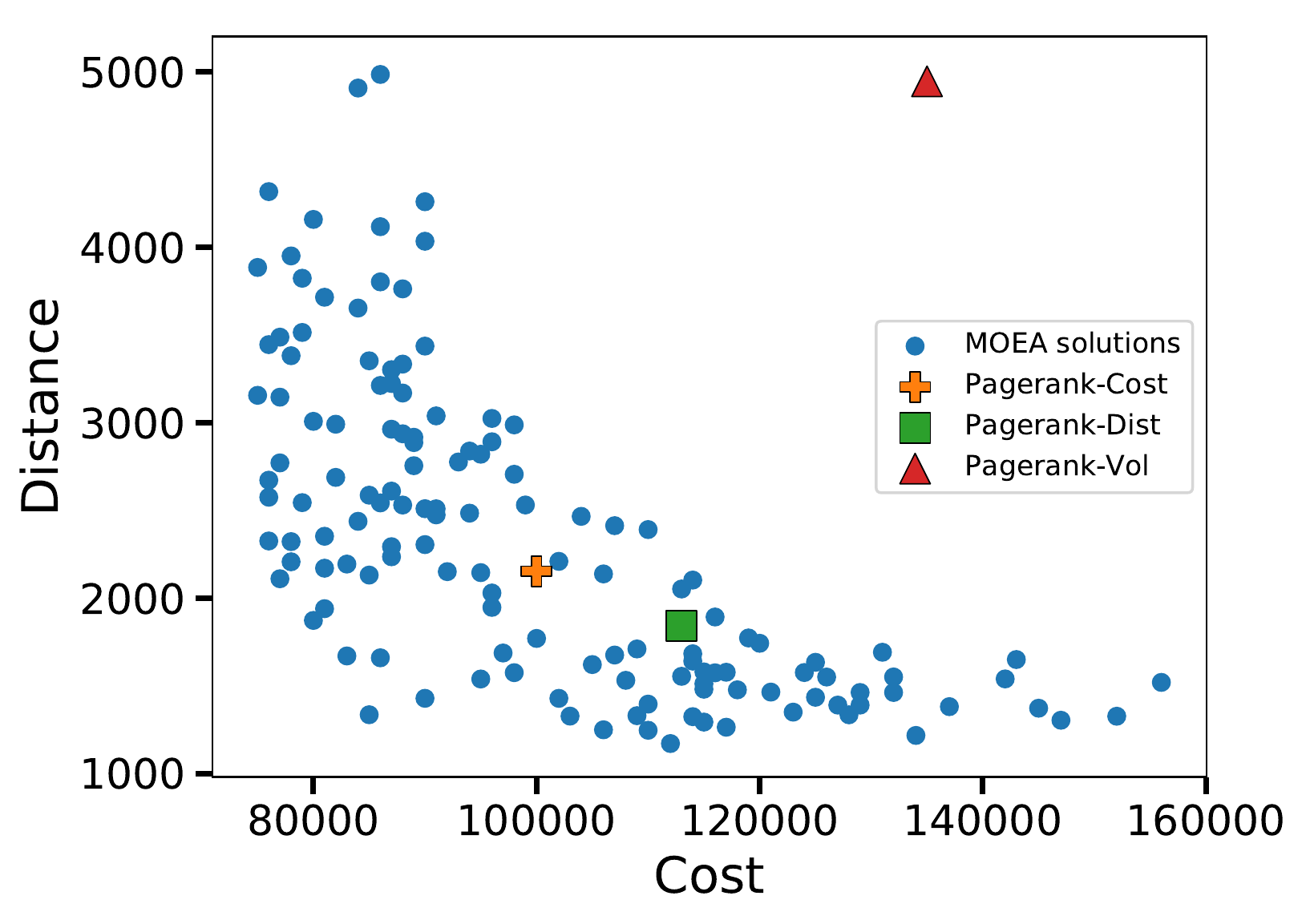} & \includegraphics[width=0.5\textwidth]{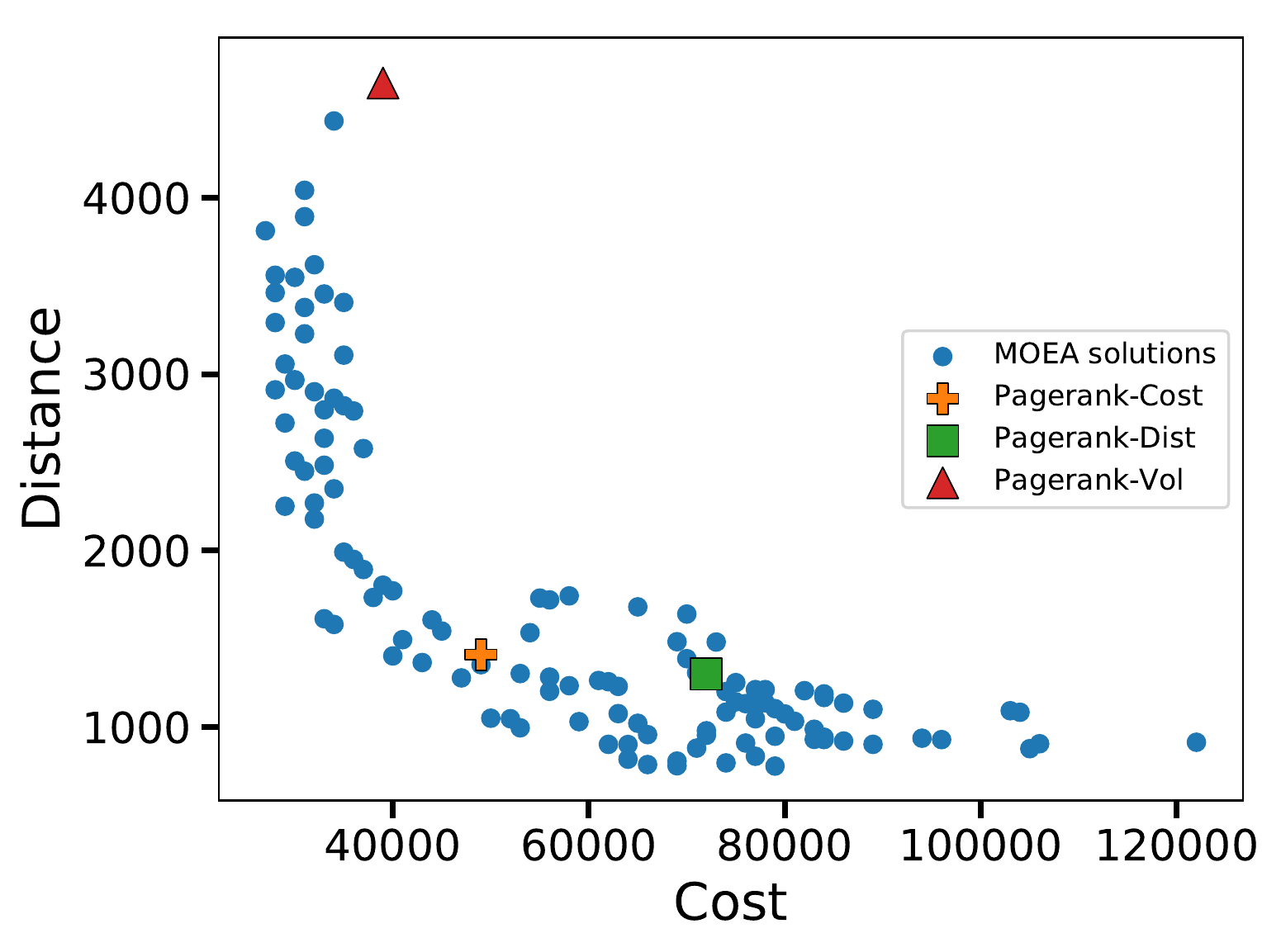}\\[-2pt]
\small{(a2) Trouville, normal demand} & \small{(b2) Villa Espa\~nola, normal demand}\\[-2pt]
\includegraphics[width=0.475\textwidth]{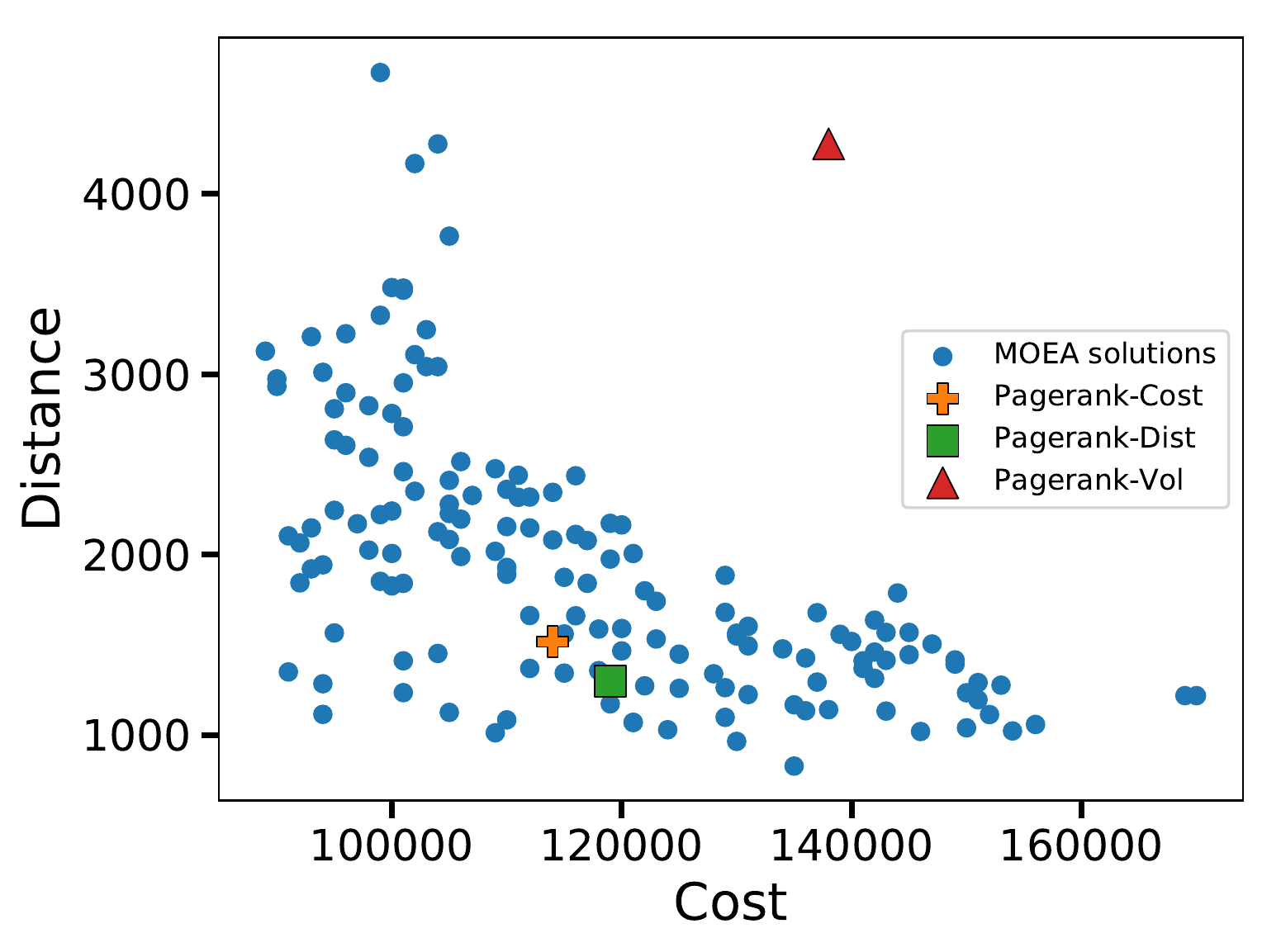} & \includegraphics[width=0.5\textwidth]{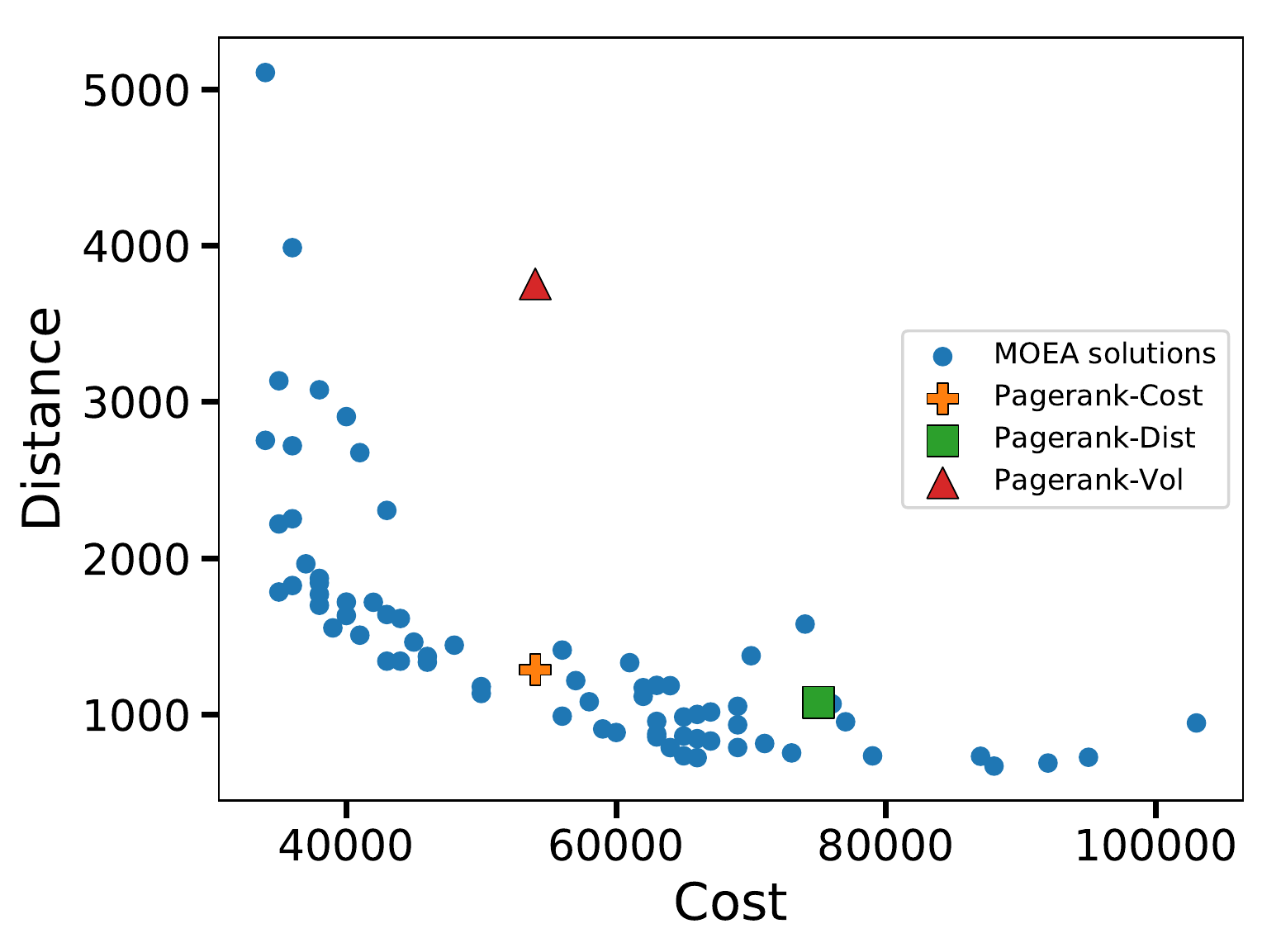}\\[-2pt]
\small{(a3) Trouville, high demand} & \small{(b3) Villa Espa\~nola, high demand}\\
\end{tabular}
\caption{2D cuts (distance/cost) of the Pareto fronts for Montevideo scenarios.}
\label{Fig:Pareto2D-MTV}
\end{figure}

\begin{figure}[!h]
\setlength{\abovecaptionskip}{3pt}
\centering    
\begin{tabular}{cc}
\includegraphics[width=0.5\textwidth]{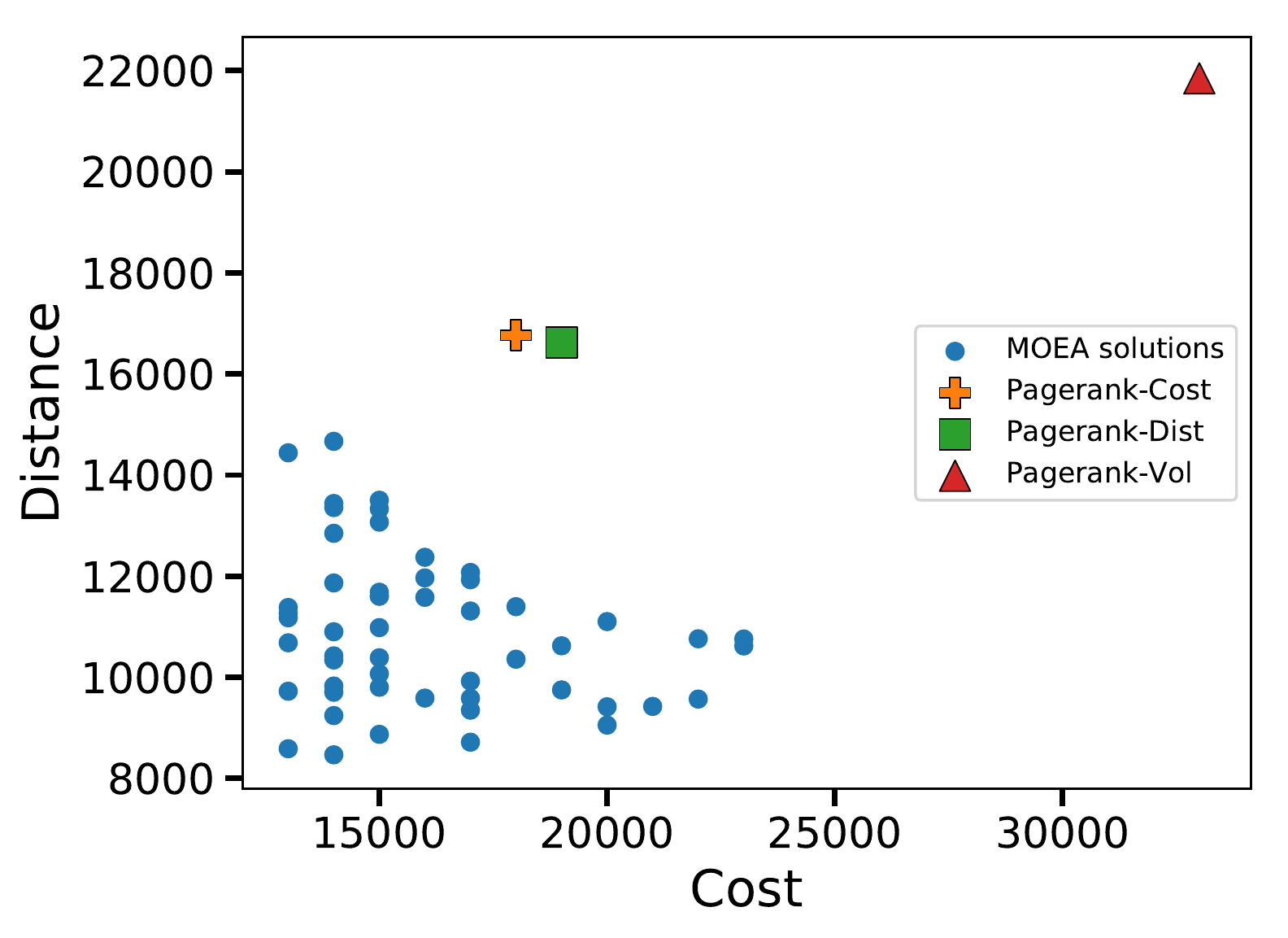} & \includegraphics[width=0.5\textwidth]{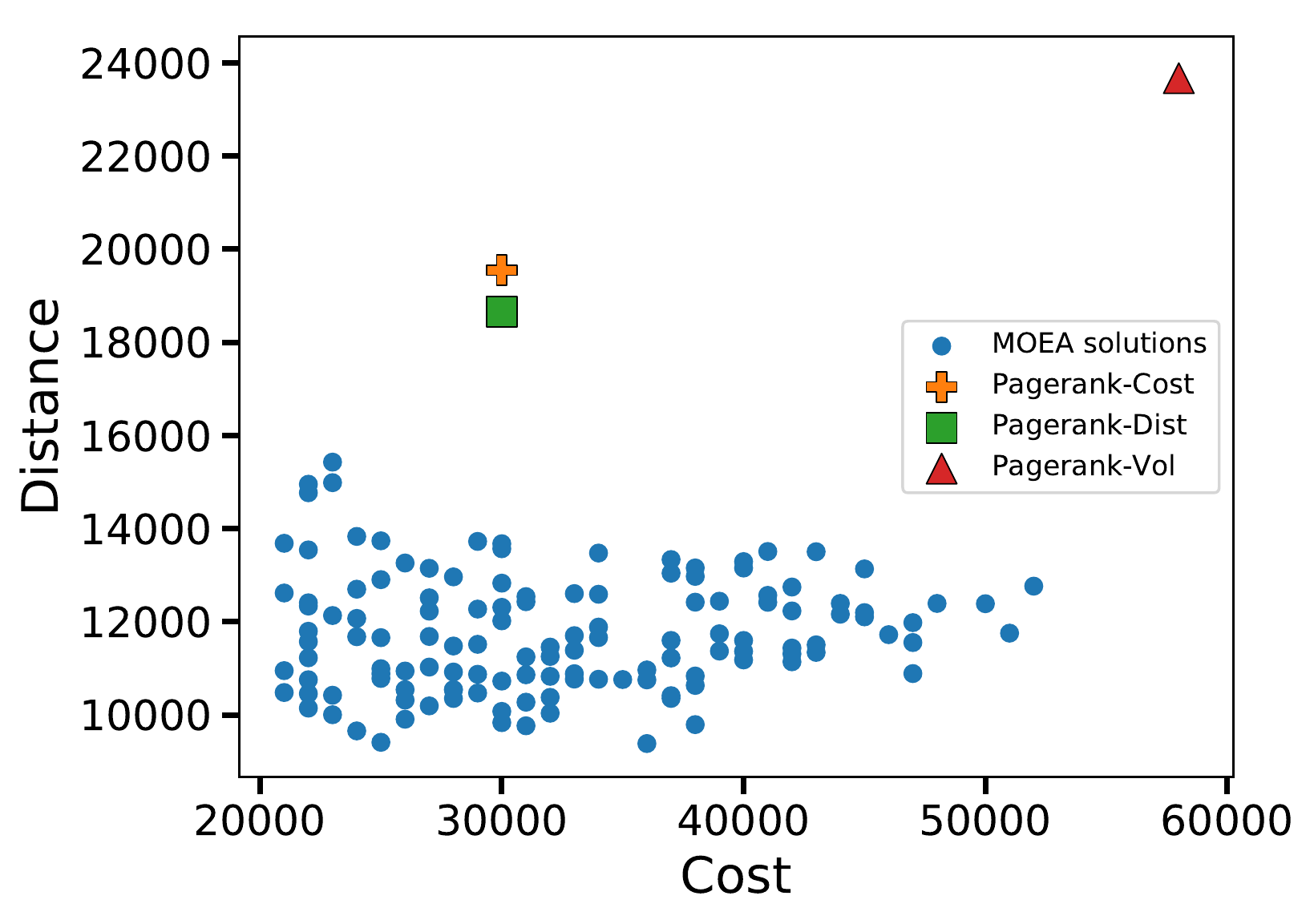}\\
\small{(a1) La Falda, low demand} & \small{(b1) Villa Mitre, low demand}\\[-1pt]
\includegraphics[width=0.5\textwidth]{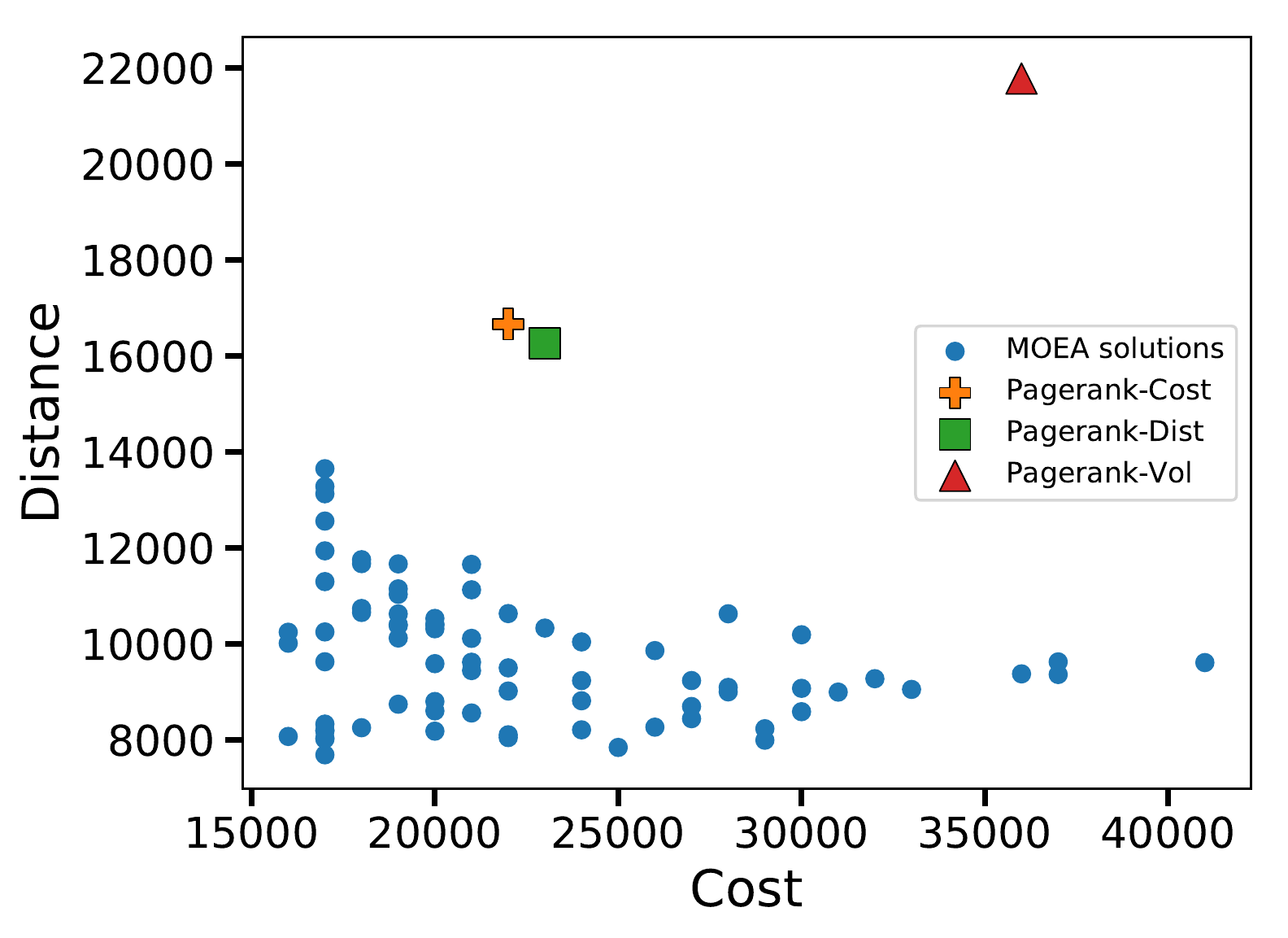} & \includegraphics[width=0.5\textwidth]{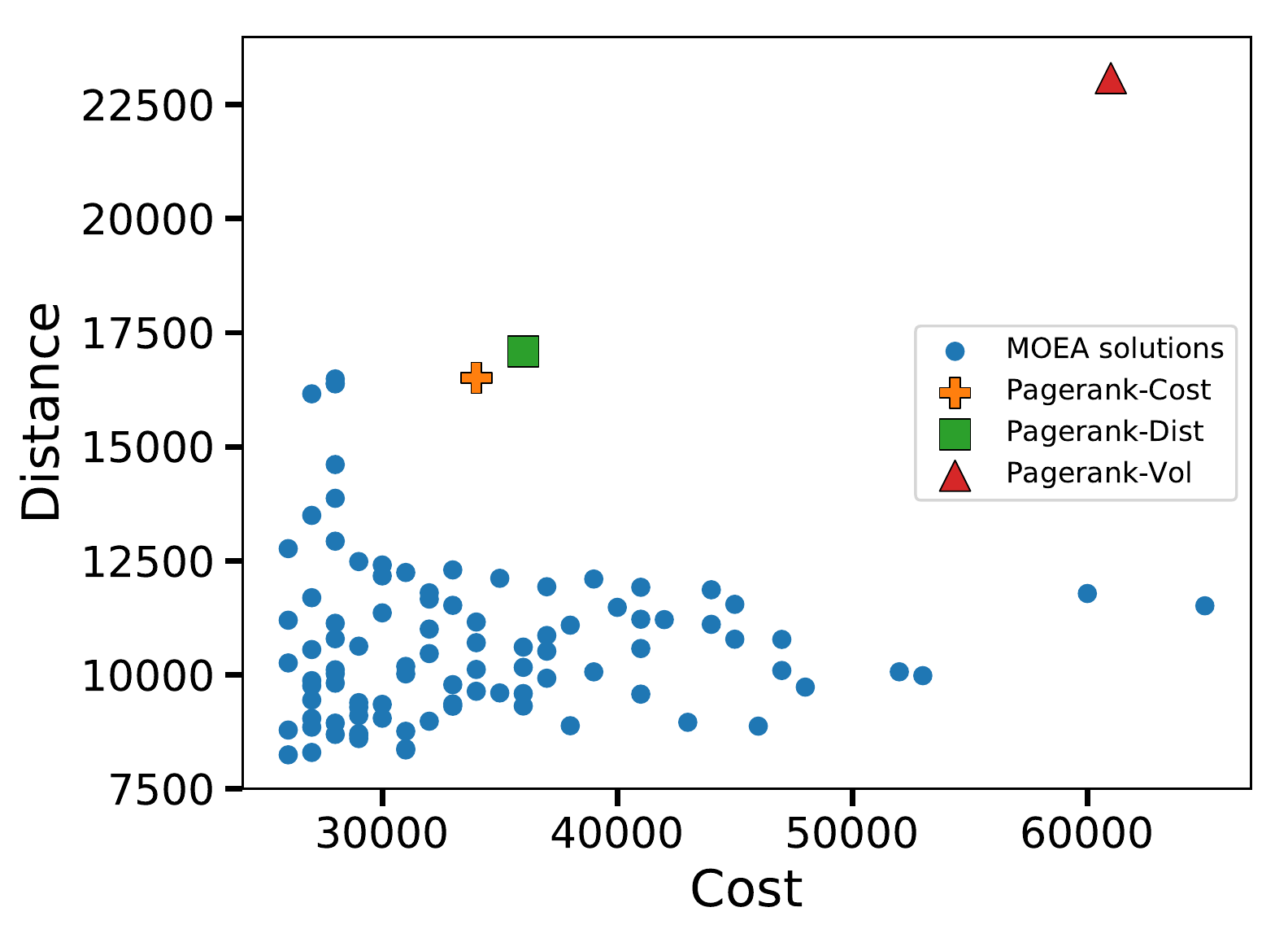}\\
\small{(a2) La Falda, normal demand} & \small{(b2) Villa Mitre, normal demand}\\[-2pt]
\includegraphics[width=0.5\textwidth]{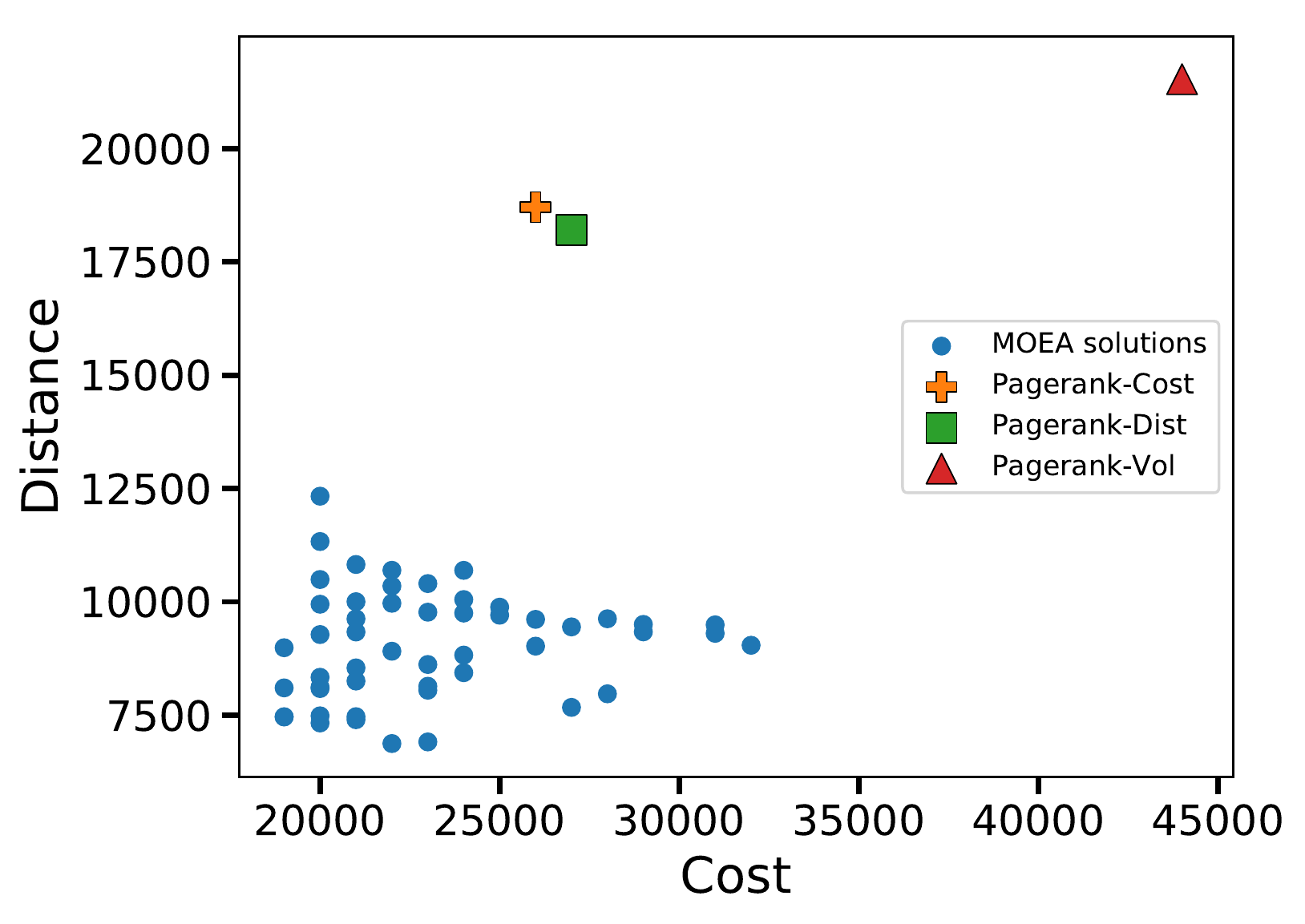} & \includegraphics[width=0.5\textwidth]{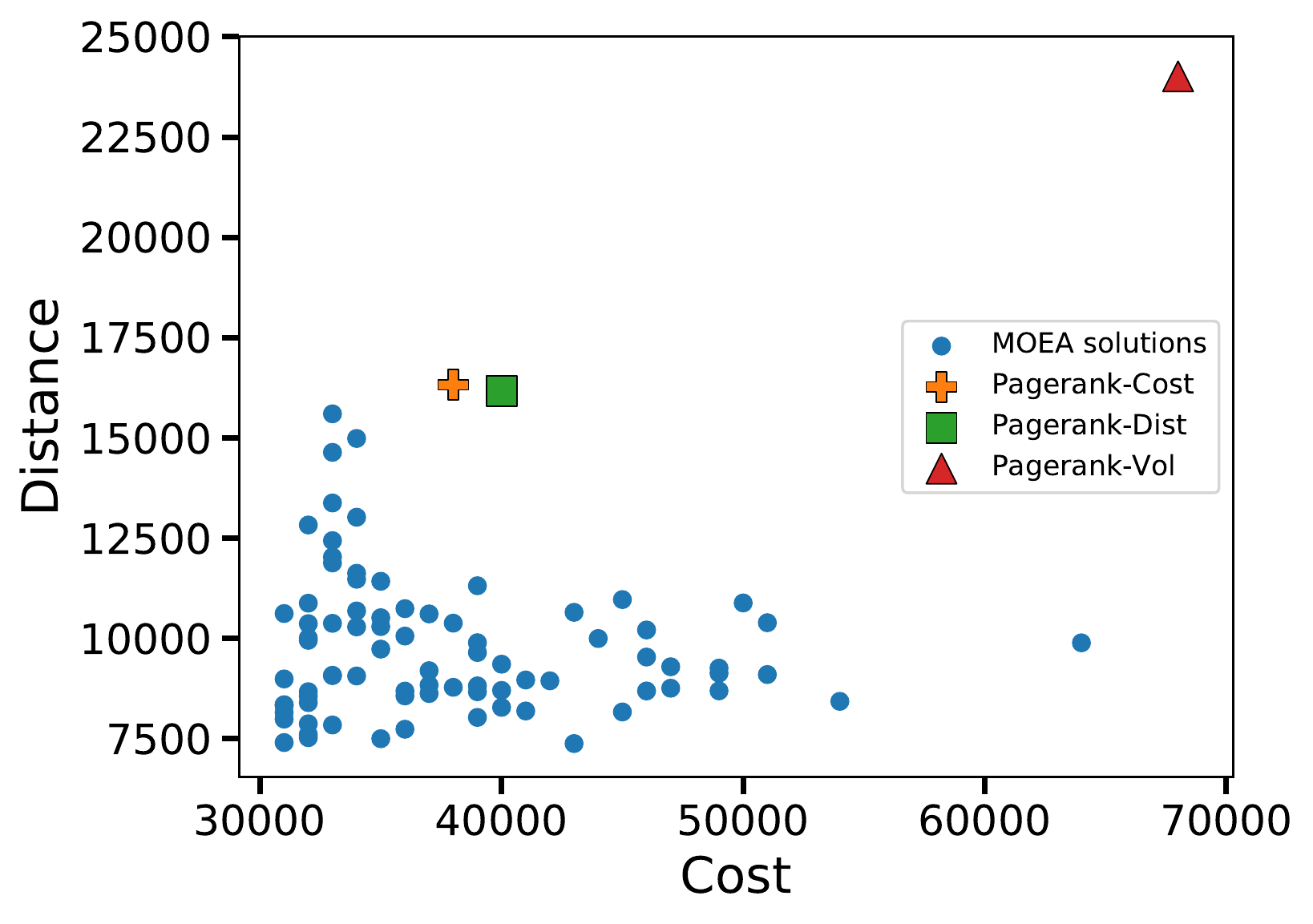}\\
\small{(a3) La Falda, high demand} & \small{(b3) Villa Mitre, high demand}\\
\end{tabular}
\caption{2D cuts (distance/cost) of the Pareto fronts for Bah\'ia Blanca scenarios.
}
\label{Fig:Pareto2D-BBCA}
\end{figure}

Fig.~\ref{Fig:Pareto2D-MTV} presents 2D cuts of the Pareto fronts obtained by {NSGA-II} and the results computed by the PageRank heuristics for Montevideo scenarios. Fig.~\ref{Fig:Pareto2D-BBCA} presents the same information for Bah\'ia Blanca scenarios. The reported results correspond to all solutions in the computed Pareto front that have a difference of up to 10\% on the volume of the total waste generated in the scenario, i.e., the installed bins receive at least 90\% of the total volume of waste generated in the scenario.


Results in Figures \ref{Fig:Pareto2D-MTV} and~\ref{Fig:Pareto2D-BBCA} demonstrate that {NSGA-II} is able to compute solutions that dominate the ones of the three PageRank algorithms for all studied scenarios, obtaining significant improvements regarding both distance and cost objectives. {NSGA-II} is also able to properly sample the Pareto front of the problem for Montevideo and Bah\'ia Blanca scenarios. 
The superiority of PageRank-Dist and PageRank-Cost over PageRank-Vol solutions is also clearly demonstrated from the reported results. This is not surprising, because PageRank-Vol does not prioritize cost or distance as an objective. Solutions computed by PageRank-Vol are generally far from the Pareto front (except for Villa Espa\~nola, normal demand scenarios). The performance of PageRank-Dist and PageRank-Cost is better in Montevideo scenarios, whereas in the Bah\'ia Blanca scenarios, PageRank results are far from the bulk of solutions in the approximated Pareto front.   
%

\subsubsection{Comparison with current GAP locations in Montevideo}

\vspace{-3mm}
Results 
were also compared with the current location of GAPs, according to real data from the government of Montevideo~\cite{MML}. The comparison is not possible for Bah\'ia Blanca, 
since no bins are currently installed in that city.

Table~\ref{Tab:comp} reports the improvements of the best compromise solution (i.e., the closest solution to the ideal vector~\cite{Deb2001}) computed by NSGA-II and the solutions of PageRank-Cost, PageRank-Dist. {MO-PageRank is not included in the comparison because of its poor solutions}. Results are reported for each scenario, considering different waste generation rates and the real GAP location and configuration in Montevideo, as of February, 2018.

\begin{table}[!h]
\setlength{\tabcolsep}{12pt}
\renewcommand{\arraystretch}{0.95}
\renewcommand{\abovecaptionskip}{3pt}
\centering
\small
\caption{Comparison of the proposed algorithms with the {real} solutions (GAP locations) in Montevideo.}
\label{Tab:comp}
\begin{tabular}{ccrrr}
\toprule
\multicolumn{5}{c}{\textit{Trouville}} \\ 
\midrule
\multirow{3}{*}{{NSGA-II}} & \textit{distance} & 84.6\% & 84.2\% & 84.2\% 	  \\ 
 & \textit{cost} & 31.4\% & 23.6\% & 10.7\% \\ 	  
 & \textit{volume} & 4.5\% & 4.0\% & 3.3\%  \\ 
 \midrule
\multirow{3}{*}{PageRank-Cost} & \textit{distance} & 84.8\% & 85.4\% & 87.6\% 	  \\ 
 & \textit{cost} & 38.2\% & 23.7\% & 13.0\% \\ 	  
 & \textit{volume} & 0.0\% & 0.0\% & 0.0\% 	 \\ 
 \midrule
\multirow{3}{*}{PageRank-Dist} & \textit{distance} & 88.6\% & 87.5\% & 89.4\% 	  \\ 
 & \textit{cost} & 24.4\% & 13.7\% & 9.2\%  \\ 	  
 & \textit{volume} & 0.0\% & 0.0\% & 0.0\%  \\ 
  \midrule
\multicolumn{5}{c}{\textit{Villa Espa\~nola}} \\ 
\midrule
\multirow{3}{*}{{NSGA-II}} & \textit{distance} & 86.0\% & 86.3\% & 87.2\% \\
& \textit{cost} & 2.3\% & -4.4\% & 4.1\% \\
 & \textit{volume} & 4.1\% & 3.3\% & 4.8\%   \\ 
 \midrule	
\multirow{3}{*}{PageRank-Cost} & \textit{distance} & 89.5\% & 90.1\% & 90.7\%  \\	
 &  \textit{cost} & 24.1\% & 15.5\% & 6.9\%   \\	
 &  \textit{volume} & 0.0\% & 0.0\% & 0.0\%   \\ 
 \midrule	
\multirow{3}{*}{PageRank-Dist} & \textit{distance} & 90.3\% & 90.9\% & 92.2\%  \\	
 & \textit{cost} & -20.7\% & -24.1\% & -29.3\% \\	
 & \textit{volume} & 0.0\% & 0.0\% & 0.0\%  \\ 
  \midrule
\midrule  
\end{tabular}
\end{table}

Regarding the distance objective, the solutions computed by the proposed approaches account for significant improvements over the current GAP location in Montevideo. For Trouville scenarios, with the current GAP location a citizen must walk
between 150.81--199.75 $m$ to dispose the waste, while in the {NSGA-II} solution the distance is reduced to 28.75--23.84m (average reduction of 84\%). For the Villa Espa\~{n}ola scenario, distances are reduced 86\% on average, from 167--188 $m$ to 21--26 $m$. The cost objective is also improved in most cases, although there are certain exceptions, such as the normal demand case in the Villa Espa\~{n}ola scenario for {NSGA-II} and the same scenario with any demand patrons for the PageRank-Dist. The PageRank algorithms collect all the available waste while {NSGA-II} leaves on average 4\% uncollected.

\subsubsection{Bins distribution}

{This subsection presents an illustrative case of the distribution of bins in each GAP for a representantive problem instance, which provides an insight on the quality of service provided to the citizens.}
   


\begin{figure}[h!]
\setlength{\abovecaptionskip}{6pt}
    \centering
    \begin{subfigure}[b]{0.99\textwidth}
        \includegraphics[width=0.99\textwidth]{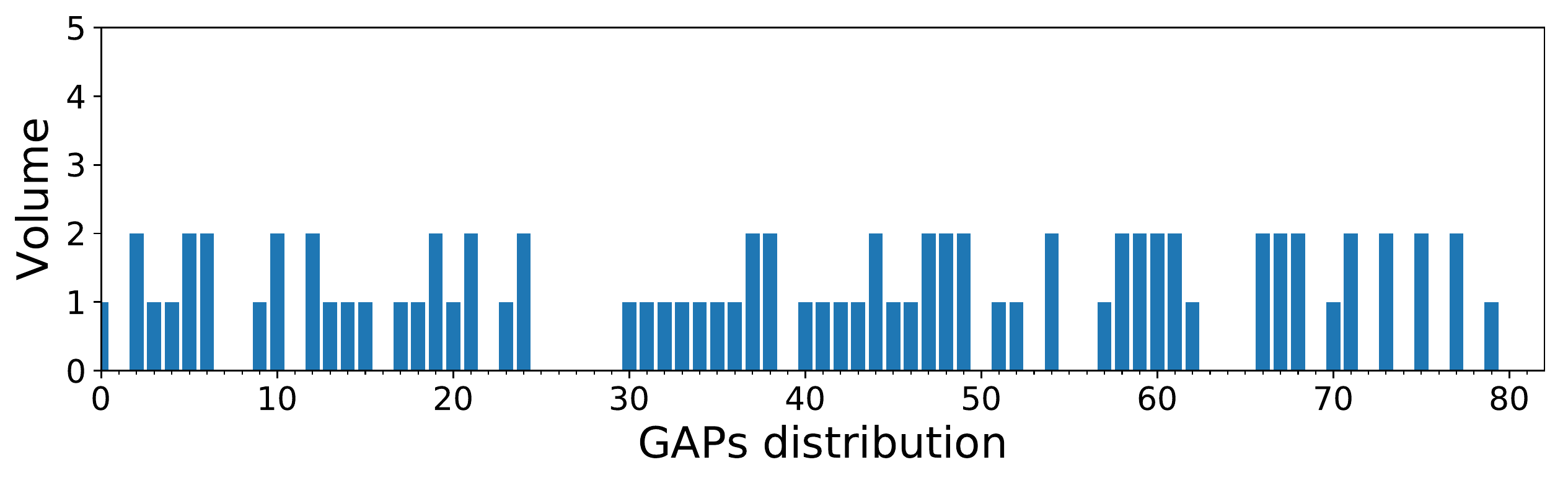}
        \caption{Trouville, normal demand, PageRank-Cost solution}
        \label{fig:hist_PR-Cost_Trouville}
   \end{subfigure}
   \begin{subfigure}[b]{0.99\textwidth}
        \includegraphics[width=0.99\textwidth]{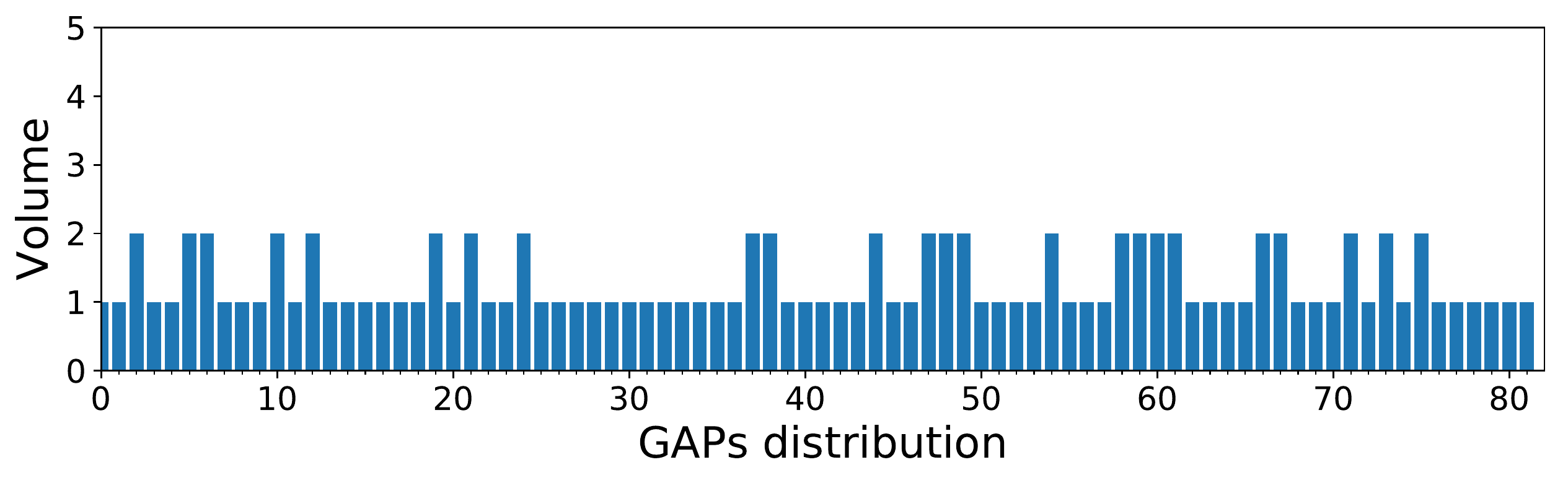}
        \caption{Trouville, normal demand, PageRank-Dist solution}
        \label{fig:hist_PR-Dist_Trouville}
   \end{subfigure}
   \begin{subfigure}[b]{0.99\textwidth}
        \includegraphics[width=0.99\textwidth]{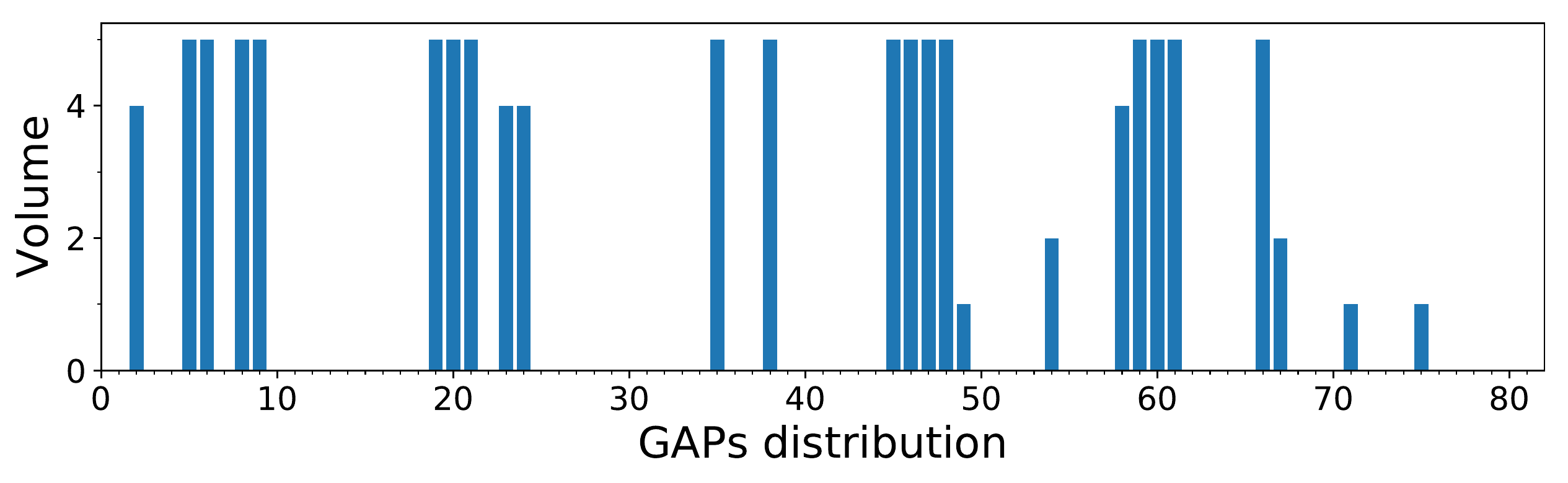}
        \caption{Trouville, normal demand, PageRank-Vol solution}
        \label{fig:hist_PR-Vol_Trouville}
   \end{subfigure}
    \begin{subfigure}[b]{0.99\textwidth}
        \includegraphics[width=.99\textwidth]{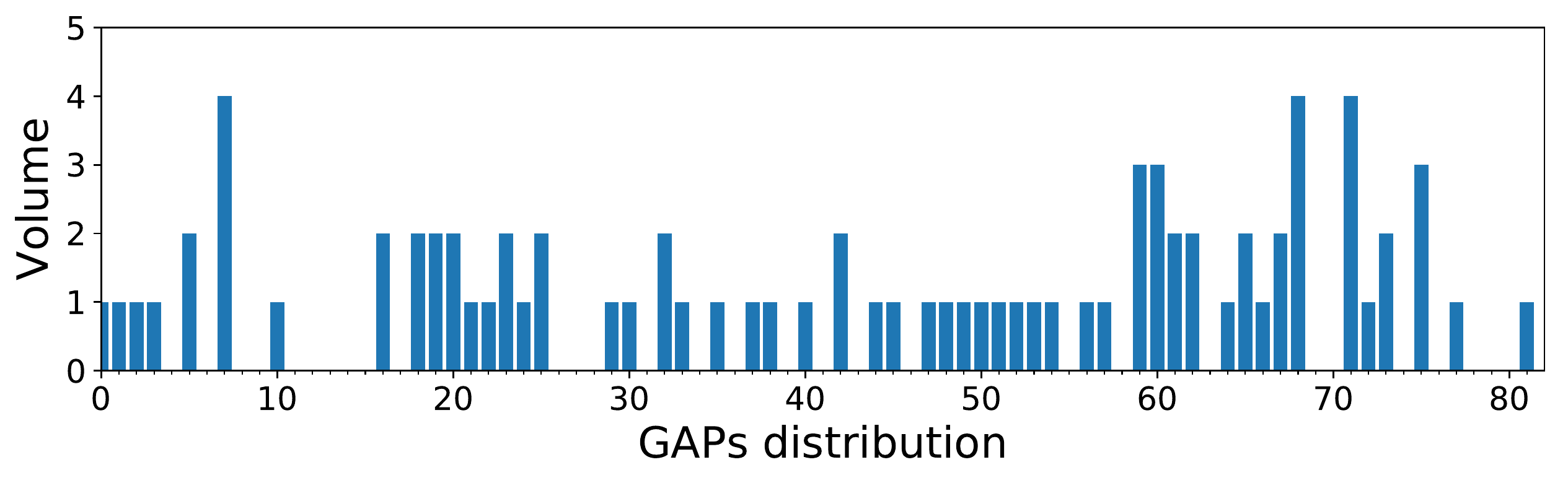}
        \caption{Trouville, normal demand, NSGA-2 solution that minimizes the installation cost}
        \label{fig:hist_MOEA_Trouvilled}
    \end{subfigure}
    \begin{subfigure}[b]{0.99\textwidth}
        \includegraphics[width=.99\textwidth]{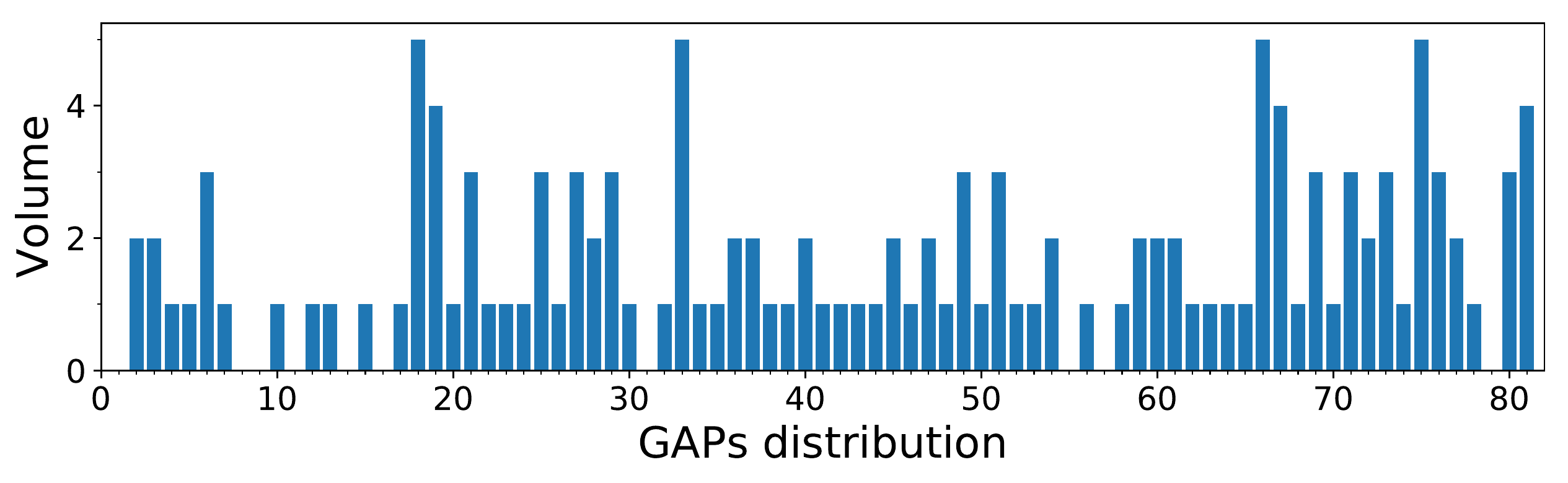}
        \caption{Trouville, normal demand, NSGA-2 solution that minimizes the walking distance}
        \label{fig:hist_MOEA_Trouvillee}
    \end{subfigure}
\caption{Installed capacity in each GAP for representative solutions of the Trouville scenario.}
\label{fig:hist_Trouville}
\end{figure}

{Fig.~\ref{fig:hist_Trouville} illustrates the distribution of the bins along the collection points and their volume in the Trouville scenario with normal demand, for the solutions computed by the single-objective PageRank methods and by NSGA-II. Again, other MO-PageRank solutions were not considered in the analysis due to their poor quality.
The figure represents the capacity (volume) of the installed configuration for each GAP (defined by its id in the x-axis).  
Two relevant NSGA-II solutions were selected for the analysis: {NSGA-II MinCost} (see Fig.~\ref{fig:hist_MOEA_Trouvilled}), which minimizes the cost objective and {NSGA-II MinDist} (see Fig.~\ref{fig:hist_MOEA_Trouvillee}), which minimizes the distance objective.}

\FloatBarrier

The analysis of results on Fig.~\ref{fig:hist_Trouville} allows concluding that PageRank-Vol computes a solution that use fewer GAPs (i.e., in several GAPs no bins are installed), but almost all the GAPs used have a remarkably large installed capacity in comparison to the others PageRank solutions. Conversely, PageRank-Dist uses all the GAPs. Despite presenting some GAPs with an installed capacity of $2 m^3$, the majority of GAPs have only one bin of type $j_1$ (with $C_{j1} = 1 m^3$). Although this type of solutions is convenient from the point of view of the generators, since it has the minimal walking distance, it has some relevant drawbacks: installation costs significantly increase and it is not efficient for collection, since every point has to be visited by the collection vehicle. PageRank-Cost computes an intermediate solution, in which a number of GAPs remain unused, and in those used GAPs the installed capacity alternates between $1 m^3$ and $2 m^3$. 

On the other hand, {NSGA-II} solutions are more balanced and provide better trade-off between the different objectives. For example, {NSGA-II}-MinDist uses a large number of GAPs, but not all of them as PageRank-Dist. To avoid using all the GAPs (and thus increasing the cost), the installed capacity of some GAPs is incremented to receive a larger amount of garbage. {NSGA-II}-MinCost follows a similar pattern but select those GAPs and configurations that allow reducing the overall cost of the installed infrastructure.

\section{Conclusions and future work}
\label{sec:conclusion}

This article introduced an optimization model for locating 
GAPs
in an urban area, aimed at maximizing the collected waste, minimizing the distance between users and 
bins, 
and minimizing the investment cost. This is a relevant problem in modern smart cities. It constitutes the initial stage in MSW reverse logistic chain, in which it has an important impact not only in the QoS provided to the citizens but also in the budget expenses of city government.

Soft computing methods, namely three PageRank greedy algorithms and {two} Multiobjective Evolutionary Algorithms, were proposed to solve the problem. In the case of the PageRank heuristics, three different {single-objective} algorithms were devised, focusing on the different objectives that were considered in the problem model. {In addition, a multiobjective variant of PageRank was designed, following a linear aggregation approach}.
The {MOEAs} applies a Pareto-based evolutionary search, allowing to compute solutions with different trade-off between the problem objectives. 

The experimental evaluation was performed on real scenarios of 
Montevideo, Uruguay, and Bah\'ia Blanca, Argentina. Three waste generation patterns were considered for each scenario (normal, low, and high demand) 
to represent the expected fluctuations of waste generation rate along the year. Results showed that {the proposed MOEAs} outperformed PageRank in all the studied scenarios regarding both distance walked to dispose the waste, which is the main QoS metric from the point of view of the citizens 
and cost objectives, the main metric considered by the city administration (installation cost of GAPs). 

{According to the analysis of multiobjective optimization metrics, NSGA-II was the best MOEA, computing slightly better solutions than SPEA-2 regarding hypervolume and also having better distribution of non-dominated solutions in the Pareto front, as confirmed by the spread results. MO-PageRank was not able to comute accurate solutions, suggesting that the linear aggregation approach is not anough to solve the problem.
Thus, NSGA-II was considered in the results comparison against single-objective PagerRank methods and it was also compared with the current GAPs location in Montevideo.}

{The comparison of the best compromise solution computed by NSGA-II with PageRank solutions showed the following results.}
In Montevideo scenarios, regarding distance, {the improvements of NSGA-II} over PageRank solutions were {16.02\%} in average, and up to 40.0\% in Villa Espa\~nola, normal demand scenario. Regarding cost, {the improvements of NSGA-II} over PageRank solutions were 7.40\% in average, and up to 20.2\%, in Trouville, high demand scenario. 
In Bah\'ia Blanca scenarios, regarding distance, {the improvements of NSGA-II} over PageRank solutions were 16.77\% in average, and up to 62.2\% in La Falda, high demand scenario. Regarding cost, {the improvements of NSGA-II} over PageRank solutions were 53.45\% in average, and up to 99.9\%, in Villa Mitre, normal and high demand scenarios.

The analysis of the Pareto fronts confirmed that {the proposed MOEAs were} able to compute accurate solutions, with different trade-off values between the problem objectives and was able to properly sample the Pareto front. {NSGA-II} solutions dominated all PageRank solutions in all but one scenario. Solutions computed by the proposed soft computing approaches also improved over the current solution applied by the authorities in the case of Montevideo. Regarding the distance objective, the best {NSGA-II} results computed allow reducing from 150m to 25m the average distance that a citizen must walk to dispose the waste, improving in 6.9\% the installation cost and maintaining the waste volume collected. This is an important result that suggests specific benefits for the city in order to apply a smart MSW system. In the case of Bah\'ia Blanca, a city that does not currently use a community bins system, the solutions can be used as a starting point by the authorities since they are considering the migration to a community bins system following the trend that has started in other important Argentinian cities.

The main lines for future work are related to extend the experimental evaluation of the proposed algorithms on other scenarios and include specific features such as the aleatory nature of the generation waste, following a more integral approach through stochastic programming. 
{Extending the algorithmic approach to consider other multiobjective algorithms is a worth line of work, too}. Furthermore, the inclusion of uncertainty in the decision making process will enhance the robustness of the solutions, so the study of algorithms for optimization under uncertainty is a meaningful idea. {Finally, from the analysis of the illustrative case of bins distribution that was performed in this paper, another research line for future work is including an objective for enhancing a uniform (or weighted-uniform) distribution of the bins along the urban area.}

\begin{acknowledgements}
We would like to thank the anonymous reviewers for their insightful comments on the paper that led us to an improvement of this work.

The work of J.~Toutouh has been
partially funded by Ministerio de Econom\'ia, Industria y Competitividad, Gobierno de Espa\~na, and European Regional Development Fund grant numbers TIN2016-81766-REDT (\url{http://cirti.es}), and TIN2017-88213-R (\url{http://6city.lcc.uma.es}). 
European Union’s Horizon 2020 research and innovation programme under the Marie Skłodowska-Curie grant agreement No 799078.
Universidad de M\'alaga, Campus Internacional de Excelencia Andaluc\'ia TECH.
\end{acknowledgements}

\bibliographystyle{spmpsci}      
\bibliography{refs}   

\end{document}